\documentclass[runningheads]{llncs}

\usepackage[T1]{fontenc}

\usepackage{graphicx}
\usepackage{comment}
\usepackage{amsmath,amssymb}
\usepackage{color}
\usepackage{url}
\usepackage{hyperref}

\usepackage{booktabs}
\usepackage{acronym}
\usepackage{multirow}
\usepackage{tabularx}
\usepackage{algorithm}
\usepackage{algpseudocode}
\usepackage{graphicx}
\usepackage{xspace}
\usepackage{xcolor}
\usepackage{tabularx}
\usepackage{etoolbox,siunitx}
\usepackage{pifont}
\usepackage{wrapfig}
\usepackage{cleveref}
\usepackage{adjustbox}
\usepackage{makecell}
\usepackage{afterpage}
\usepackage{mathtools}
\usepackage{wrapfig}
\usepackage{lipsum}
\usepackage{setspace}
\usepackage{tocloft}
\usepackage{minitoc}
\usepackage[absolute,overlay]{textpos}

\definecolor{ourgreen}{HTML}{6bc958}
\definecolor{ourred}{HTML}{be0000}

\acrodef{method}[CoProU-VO]{Combined Projected Uncertainty VO}
\acrodef{VO}[VO]{Visual Odometry}
\acrodef{SSIM}[SSIM]{Structural Similarity Index}
\acrodef{projection}[CoProU]{Combined Projected Uncertainty}

\newif\ifreview
\reviewfalse

\ifreview
	\usepackage{lineno}

	\linenumbers
\fi

\begin{document}


\def\SubNumber{27}

\def\GCPRTrack{Main Track}

\title{CoProU-VO: Combining Projected Uncertainty for End-to-End Unsupervised Monocular Visual Odometry}

\ifreview
\else
	\titlerunning{CoProU-VO}

	\author{Jingchao Xie\inst{^*,1,3} \quad
	Oussema Dhaouadi\inst{^*,1,2,3,}$^{\dagger}$ \quad
	Weirong Chen\inst{1,3}\\ 
        Johannes Meier\inst{1,2,3} \quad
        Jacques Kaiser\inst{2} \quad
        Daniel Cremers\inst{1,3}
}
	
	\authorrunning{Xie et al.}
	
	\institute{\quad $^1$TU Munich \quad $^2$\href{https://www.deepscenario.com}{DeepScenario} \quad $^3$ \href{https://mcml.ai}{Munich Center for Machine Learning}}
	
\fi

\setcounter{footnote}{0}
\renewcommand{\thefootnote}{}

\footnotetext{*Shared first authorship, $\dagger$ Corresponding author.}
\footnotetext{TUM: {\{jingchao.xie, oussema.dhaouadi, weirong.chen, j.meier, cremers\}@tum.de}}
\footnotetext{DeepScenario: \{firstname\}@deepscenario.com}
\renewcommand{\thefootnote}{\arabic{footnote}} 

\begin{textblock*}{\textwidth}(4.8cm,24cm)
  \noindent
  \fbox{%
    \parbox{\dimexpr\textwidth-2\fboxsep-2\fboxrule}{%
      \footnotesize
      To appear in Proceedings of the 47th DAGM German Conference on Pattern Recognition (GCPR), 2025. The final publication will be available through Springer.
    }%
  }
\end{textblock*}

\maketitle              

\begin{abstract}

Visual Odometry (VO) is fundamental to autonomous navigation, robotics, and augmented reality, with unsupervised approaches eliminating the need for expensive ground-truth labels. However, these methods struggle when dynamic objects violate the static scene assumption, leading to erroneous pose estimations. We tackle this problem by uncertainty modeling, which is a commonly used technique that creates robust masks to filter out dynamic objects and occlusions without requiring explicit motion segmentation. Traditional uncertainty modeling considers only single-frame information, overlooking the uncertainties across consecutive frames. Our key insight is that uncertainty must be propagated and combined across temporal frames to effectively identify unreliable regions, particularly in dynamic scenes. To address this challenge, we introduce Combined Projected Uncertainty VO (CoProU-VO), a novel end-to-end approach that combines target frame uncertainty with projected reference frame uncertainty using a principled probabilistic formulation. Built upon vision transformer backbones, our model simultaneously learns depth, uncertainty estimation, and camera poses. Consequently, experiments on the KITTI and nuScenes datasets demonstrate significant improvements over previous unsupervised monocular end-to-end two-frame-based methods and exhibit strong performance in challenging highway scenes where other approaches often fail. Additionally, comprehensive ablation studies validate the effectiveness of cross-frame uncertainty propagation. The code is publicly available at \url{jchao-xie.github.io/CoProU}.

\end{abstract} 


\section{Introduction}
\label{sec:introduction}

\begin{figure}[th]
    \centering
    \includegraphics[width=1\textwidth]{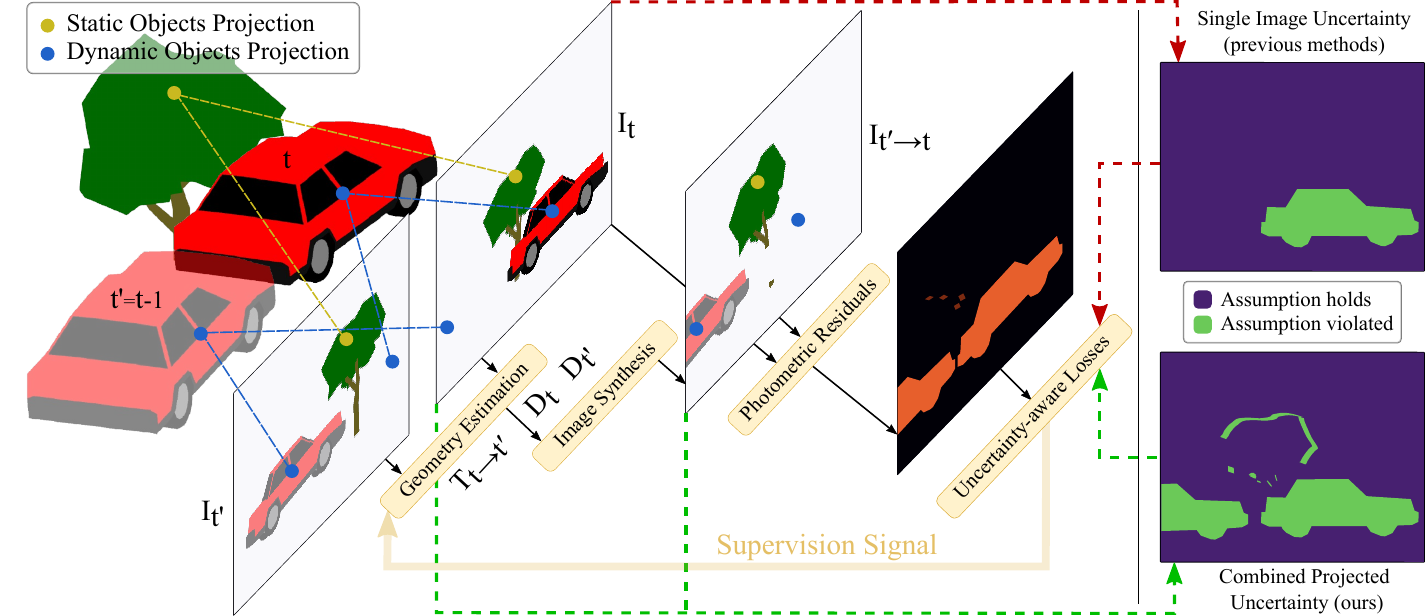}
    \caption{\textbf{Our Proposed Combined Projected Uncertainty.}
Unsupervised monocular VO typically synthesizes a target image $I_{t}$ from its reference image $I_{t'}$ using the relative pose $T_{t \to t'}$ and depth $D_t$. The error between the target image $I_{t}$ and the synthesized image $I_{t' \to t}$ can serve as a supervision signal for geometry estimation, based on the static scene assumption. However, this assumption is violated by dynamic objects and occlusions. Previous methods (\textcolor{ourred}{red flow}) attempt to mask out these elements by predicting uncertainty solely from the target image, which overlooks the uncertainty arising from the reference image. In contrast, our proposed combined projected uncertainty (\textcolor{ourgreen}{green flow}) considers the uncertainty from both the target and reference images, and robustly masks out all areas violating the static scene assumption.}
    \label{fig:uncertainty_comparison}
\end{figure}

\ac{VO} has emerged as a fundamental component in numerous applications including autonomous driving~\cite{geiger2013vision}, augmented reality~\cite{chen2023deep}, and robotics~\cite{yousif2015overview} due to its ability to estimate camera motion using only visual inputs. Despite significant advances, state-of-the-art \ac{VO} methods still face one or more of the following limitations: the requirement for labeled real-world data, poor real-time performance, and inaccurate motion estimation in dynamic scenes.

While supervised approaches~\cite{wang2024dust3r,leroy2024grounding,wang2025vggt,zhang2024monst3r} have demonstrated impressive performance in multi-view 3D scene reconstruction and camera pose recovery, with Monst3r~\cite{zhang2024monst3r} and VGGT~\cite{wang2025vggt} showing strong results even in dynamic scenes, their reliance on ground-truth annotations inherently limits scalability and generalization to diverse environments. Similarly, although unsupervised methods with supervised pretraining, such as CasualSAM~\cite{zhang2022structure} and AnyCAM~\cite{wimbauer2025anycam}, demonstrate promising results, they depend heavily on predictions from pre-trained optical flow and metric depth foundation models, which are themselves trained on labeled real-world data, thereby constraining the generalization of these methods, particularly in long-tail scenarios. 

Fully unsupervised \ac{VO}, which eliminates the need for annotated real-world data, can be broadly categorized into three categories: classical geometry-based methods~\cite{mur2015orb,mur2017orb,campos2021orb}, end-to-end methods~\cite{bian2021ijcv,zhou2017unsupervised,monodepth2}, and hybrid methods~\cite{yang2020d3vo,wang2024self,dai2022self,bangunharcana2023dualrefine}, which combine elements of the first two. Classical methods often fail in highly dynamic scenes, while hybrid methods typically involve computationally intensive optimization, limiting their real-time applicability and scalability to large-scale data. Therefore, in this work, we focus on fully end-to-end approaches.

SC-Depth~\cite{bian2021ijcv} is a representative fully end-to-end, two-frame-based unsupervised monocular \ac{VO} approach and serves as our baseline. As illustrated in~\Cref{fig:uncertainty_comparison}, given a target image and a reference image, the method synthesizes the target image from the reference image. Under the static scene assumption, the photometric error between the target and synthesized image is used as the supervisory signal.
It employs DepthNet and PoseNet to estimate depth and relative camera poses between consecutive frames, and introduces a geometric consistency loss to penalize depth inconsistencies, along with a self-discovered mask to filter out moving objects.
However, the method overlooks other violations of the static scene assumption, such as non-Lambertian surfaces. Furthermore, the mask based solely on depth error lacks robustness in complex scenes.

To improve robustness in \ac{VO} systems, uncertainty modeling has emerged as a powerful technique. Recent works like D3VO~\cite{yang2020d3vo} demonstrate that modeling heteroscedastic aleatoric uncertainty~\cite{kendall2017uncertainties} as a Laplacian probability distribution effectively filters out problematic regions including object boundaries, moving objects, and highly reflective surfaces. Similarly, AnyCAM~\cite{wimbauer2025anycam} and KPDepth-VO~\cite{wang2024self} incorporate uncertainty to enhance robustness against inconsistencies. However, all the unsupervised methods have a fundamental limitation by incorporating uncertainty solely from the target image when filtering unreliable regions.

To overcome this limitation, we present \ac{method}, a novel visual odometry approach that robustly handles regions violating the static scene assumption. Our method incorporates a lightweight PoseNet trained through robust geometric constraints to maintain real-time performance and independently used at test time. We leverage pre-trained features from DepthAnything~\cite{yang2024depth,yang2024depthv2}, which is pre-trained without labeled real-world data, for joint depth and uncertainty prediction from a single image. To address violations of the static scene assumption, we project the uncertainty of the reference frame into the target frame and merge it with the native uncertainty of the target frame, which proves more robust than single-image uncertainty estimation, as demonstrated in~\Cref{fig:uncertainty_comparison}. This cross-frame uncertainty propagation establishes robust gradient flow between uncertainty and pose estimation, enhancing projection awareness and significantly improving robustness to dynamic objects and other consistency violations.

Our contributions are twofold: (1) we propose a novel projection-aware uncertainty mechanism called \textit{\ac{projection}}, which integrates uncertainties from both the target and reference frames to robustly mask regions that violate the static scene assumption; and (2) based on this mechanism, we build \textit{\ac{method}}, which incorporates pre-trained features from depth foundation models in an end-to-end unsupervised framework, while maintaining real-time performance via a lightweight PoseNet during inference.

\section{Related Work}
\label{sec:related_work}

\subsection{Unsupervised Monocular Visual Odometry}

SfMLearner~\cite{zhou2017unsupervised} was the first method to achieve fully unsupervised learning of depth and ego-motion in an end-to-end manner using only monocular video sequences. Despite this advancement, SfMLearner~\cite{zhou2017unsupervised} faces two significant limitations: inability to provide pose estimation with consistent global scale and lack of robustness in dynamic scenarios.

Numerous methods have since built upon this foundation to address these shortcomings. SC-Depth~\cite{bian2021ijcv} introduced a geometric consistency loss to enforce alignment between predicted and synthesized depths, encouraging scale consistent depth predictions. MonoDepth2~\cite{monodepth2} employed a minimum reprojection loss to handle occlusions more effectively. 

DynamicDepth~\cite{feng2022disentangling} proposed occlusion-aware training via self-supervised cycle-consistent learning, relying on semantic segmentation to detect dynamic objects. In contrast, our method effectively masks dynamic objects and occluded regions using a principled probabilistic formulation, without external segmentation.

Hybrid methods~\cite{yang2020d3vo,wang2024self,zhan2020visual,dai2022self} achieve promising results by combining end-to-end learning with classical geometric algorithms. D3VO~\cite{yang2020d3vo} integrates predicted depth, pose, and uncertainty into a windowed sparse photometric bundle adjustment framework based on Direct Sparse Odometry (DSO)~\cite{engel2017direct}. AnyCam~\cite{wimbauer2025anycam} demonstrates camera tracking using pre-trained foundation models without prior knowledge of camera intrinsics. However, these methods often rely on computationally intensive non-linear optimization at inference, limiting their practical applicability. In contrast, \ac{method} uses a straightforward single-pass inference approach that maintains accuracy while being computationally efficient.

\subsection{3D Vision Foundation Models}

The adoption of Vision Transformers~\cite{dosovitskiy2020image} in 2D vision has led to the development of several foundation models for 3D vision~\cite{wang2024dust3r,leroy2024grounding,yang2024depth,yang2024depthv2,wang2025vggt}. DUSt3R~\cite{wang2024dust3r} and MASt3R~\cite{leroy2024grounding} directly estimate aligned dense point clouds from image pairs without requiring camera parameters, enabling multi-view geometry reasoning. 

VGGT~\cite{wang2025vggt}, a Visual Geometry Grounded Transformer inspired by CLIP~\cite{radford2021learning} and DINO~\cite{caron2021emerging,oquab2024dinov2,darcet2023vision}, has shown promising results for 3D scene prediction through training on large-scale annotated data. While these supervised approaches achieve good performance, they remain limited by their dependence on labeled data.

DepthAnythingV2~\cite{yang2024depthv2}, a foundation model for monocular depth estimation, addresses this limitation through semi-supervised training. Its teacher model is initially trained on synthetic depth maps, followed by a student model trained on unlabeled real-world data using knowledge distillation~\cite{hinton2015distilling}. This eliminates the need for labeled real-world data while maintaining performance. Given its capabilities in monocular depth estimation, we adopt the DINOv2~\cite{oquab2024dinov2}-based encoder from DepthAnythingV2~\cite{yang2024depthv2} as the backbone for \ac{method}, enabling depth and uncertainty estimation without labeled real-world data.

\subsection{Uncertainty Estimation}

Uncertainty in unsupervised \ac{VO} can be categorized into two primary types: photometric and depth uncertainties. Photometric uncertainty emerges from violations of the static scene assumption, particularly due to non-Lambertian surfaces, moving objects, and occlusions that break the brightness constancy assumption. Depth uncertainty primarily manifests itself in low-texture regions or at large distances, where significant depth estimation errors produce negligible changes in the photometric reconstruction loss, making these areas inherently difficult to constrain through photometric supervision alone.

Klodt et al.~\cite{klodt2018supervising} extended SfMLearner~\cite{zhou2017unsupervised}, using a probabilistic model to estimate photometric, depth, and pose uncertainty. D3VO~\cite{yang2020d3vo} addresses heteroscedastic aleatoric uncertainty~\cite{kendall2017uncertainties} by modeling photometric loss as a Laplacian distribution, identifying unreliable regions during training.

For depth uncertainty, recent works~\cite{poggi2020uncertainty,dikov2022variational,wang2024self,marsal2024monoprob} show significant progress. 

VDN~\cite{dikov2022variational} proposed modeling depth uncertainty with a Variational Depth Network, while KPDepth-VO~\cite{wang2024self} introduced photometric-sensitive depth uncertainty to improve scale recovery. 

In this work, we focus on photometric uncertainty. Existing unsupervised \ac{VO} methods use only the uncertainty of the target image to filter unreliable regions. However, photometric loss stems from target and reference images. We tackle this by introducing the \ac{method} that integrates uncertainty information from both frames to better identify regions that violate the assumption of static scene in dynamic environments.


\section{Method}
 
\begin{figure}[t]
  \centering
  \includegraphics[width=\textwidth]{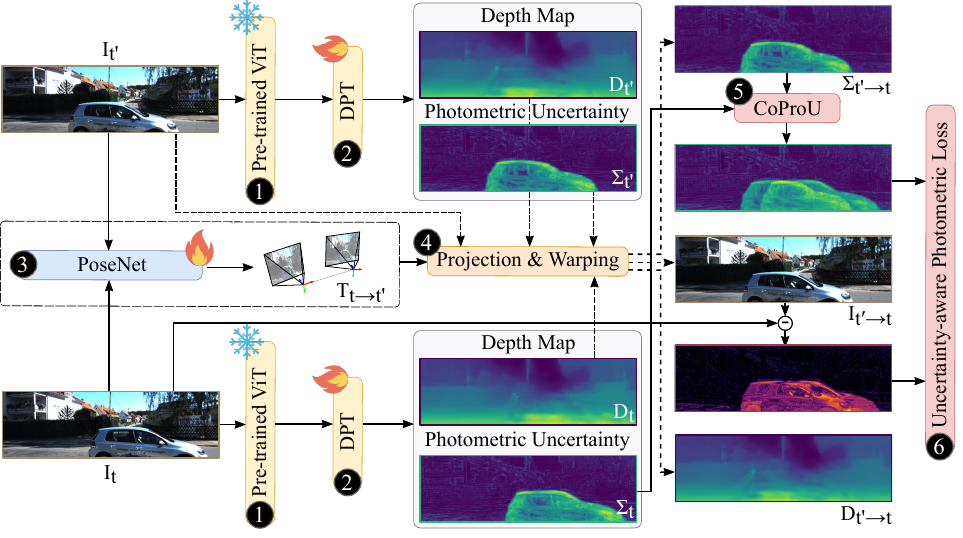}
  \caption{\textbf{Overview of Our \ac{method} Approach}. Given two consecutive frames (target $I_t$ and reference $I_{t'}$), (1) features are extracted using a pre-trained vision transformer backbone, (2) depth maps and uncertainty estimates are produced through a decoder network for both frames, (3) relative camera pose is predicted by a PoseNet module, (4) projection and warping operations synthesize views between frames, and (5) our novel \acs{projection} module integrates uncertainty information from both target and reference frames, which is used to (6) compute the uncertainty-aware loss.}
  \label{fig:pipeline}
\end{figure}

Following~\cite{zhou2017unsupervised}, given a target image $I_{t}$ at time $t$ and a reference image $I_{t'}$ at time~$t' \in \{t-1, t+1\}$, our goal is to synthesize the target image from the reference image using geometry estimation in the form of depth and relative camera pose. The photometric error between $I_{t}$ and the synthesized image $I_{t'\rightarrow t}$ serves as the unsupervised training signal in our framework, illustrated in \Cref{fig:pipeline}. We describe the image synthesis process in~\Cref{sec:image_synthesis} and formulate our novel combined projected uncertainty in~\Cref{sec:combined_projected_uncertainty}, which is the key innovation enabling robust handling of dynamic objects and occlusions. Finally, we detail how this uncertainty is integrated into our overall training objective in~\Cref{sec:losses}.

\subsection{Image Synthesis}
\label{sec:image_synthesis}

\subsubsection{Depth and pose estimation} We independently pass both target and reference images through a frozen, pre-trained vision transformer
to extract image features. These features are decoded by a randomly initialized DPT layer~\cite{ranftl2021vision} to estimate target depth $D_t$, reference depth $D_{t'}$, target photometric uncertainty $\Sigma_{t}$, and reference photometric uncertainty $\Sigma_{t'}$. To estimate the relative pose ${T}_{t \rightarrow t'} \in SE(3)$ from the target view to the reference view, we concatenate the target and reference images and process them through a modified ResNet-18 PoseNet~\cite{he2016deep}. 

\subsubsection{Projection and warping} 
Let $p_{t} \in \mathbb{R}^{3}$ denote homogeneous pixel coordinates in the target view and $K \in \mathbb{R}^{3 \times 3}$ be the camera intrinsics matrix. The projected coordinates of $p_t$ onto the reference view are:
\begin{equation}
p_{t \rightarrow t'} \sim K \, T_{t \rightarrow t'} \, D_t(p_t) \, K^{-1} \, p_t \, .
\end{equation}
For brevity, the conversion from/to homogeneous coordinates is omitted in the remainder of this paper. We employ differentiable bilinear sampling~\cite{jaderberg2015spatial,zhou2017unsupervised} to obtain the synthesized image $I_{t' \rightarrow t}$, the synthesized depth $D_{t' \rightarrow t}$, and the synthesized uncertainty $\Sigma_{t' \rightarrow t}$. The synthesized image intensity at a pixel $p_t$ is computed as:
\begin{equation}
I_{t' \rightarrow t}(p_t) = \sum_{i \in \{\text{tl}, \text{tr}, \text{bl}, \text{br}\}} w^{i} I_{t'}(p_{t \rightarrow t'}^i) \, .
\end{equation}

Here, $p_{t \rightarrow t'}^i$ with $i \in \{\text{tl}, \text{tr}, \text{bl}, \text{br}\}$ denotes the four neighboring pixels of the continuous projected coordinate $p_{t \rightarrow t'}$, and $w^i$ are spatial weights summing to 1. Similarly, $D_{t' \rightarrow t}$ and $\Sigma_{t' \rightarrow t}$ are derived from $D_{t'}$ and $\Sigma_{t'}$, respectively.

\subsection{Combined Projected Uncertainty}
\label{sec:combined_projected_uncertainty}

\subsubsection{Photometric residual} Following previous work~\cite{yang2020d3vo,monodepth2,bian2021ijcv,wang2024self}, we define the photometric residual between the target image $I_t$ and the synthesized image $I_{t' \rightarrow t}$ at pixel $p_t$ as:

\begin{adjustbox}{width=1\linewidth}
\begin{minipage}{1.13\linewidth}
\begin{equation}
\begin{split}
r\bigl(I_t(p_t), I_{t' \rightarrow t}(p_t)\bigr) = & 
\frac{\alpha}{2}\Bigl(1-\operatorname{SSIM}\!\bigl(I_t(p_t), I_{t' \rightarrow t}(p_t)\bigr)\Bigr) + (1-\alpha)\,\bigl\lVert I_t(p_t) - I_{t' \rightarrow t}(p_t)\bigr\rVert_{1}
\end{split} \, ,
\label{eq:photometric}
\end{equation}
\end{minipage}
\end{adjustbox}

\noindent where $\alpha \in [0,1]$ balances the \ac{SSIM} and $L_1$ components. To handle illumination changes and other noise sources, we incorporate the \ac{SSIM}~\cite{wang2004image}, as it captures luminance, contrast, and structural similarity.

\subsubsection{Uncertainty formulation}
In supervised methods, heteroscedastic aleatoric uncertainty in data is modeled using a Laplacian distribution, while assuming no uncertainty in the prediction according to~\cite{kendall2017uncertainties}. Existing unsupervised \ac{VO} methods, such as D3VO~\cite{yang2020d3vo}, naively adopt this single uncertainty mechanism into unsupervised \ac{VO}.

However, in unsupervised \ac{VO}, the prediction (synthesized image  $I_{t'\rightarrow t}$) also contains noise from the reference image $I_{ t'}$. To capture this, we model each pixel intensity as an independent Laplace random variable: $I_t(p_t) \sim \mathrm{Laplace}\bigl(\mu_1,\Sigma_t(p_t)\bigr)$ and $I_{t'\rightarrow t}(p_t) \sim \mathrm{Laplace}\bigl(\mu_2,\Sigma_{t'\rightarrow t}(p_t)\bigr)$, where $\mu_1$ and $\mu_2$ represent the underlying true pixel intensities.

We approximate the likelihood of the photometric residual by accounting for both uncertainty sources:

\begin{adjustbox}{width=1\linewidth}
\begin{minipage}{1.12\linewidth}
\begin{equation}
\begin{split}
p\Bigl(
  r\bigl(I_t(p_t), I_{t' \rightarrow t}(p_t)\bigr)
  \,\Big|\, \Sigma_t(p_t), \Sigma_{t' \rightarrow t}(p_t)\, 
\Bigr)
= \frac{1}{2\sigma_{\text{eff}}(p_t)}
\exp\left(
  -\frac{r\bigl(I_t(p_t), I_{t' \rightarrow t}(p_t)\bigr)}
        {\sigma_{\text{eff}}(p_t)} \
\right)
\end{split} \, ,
\label{eq:laplace_photometric}
\end{equation}
\end{minipage}
\end{adjustbox}

\noindent with negative log-likelihood:

\begin{adjustbox}{width=1\linewidth}
\begin{minipage}{1.24\linewidth}
\begin{equation}
\begin{split}
- \log p\Bigl(
  r\bigl(I_t(p_t), I_{t' \rightarrow t}(p_t)\bigr)
  \,\Big|\, \Sigma_t(p_t), \Sigma_{t' \rightarrow t}(p_t)
\Bigr)
= \frac{r\bigl(I_t(p_t), I_{t' \rightarrow t}(p_t)\bigr)}
         {\sigma_{\text{eff}}(p_t)}
\;+\; \log\sigma_{\text{eff}}(p_t)
\;+\; \text{const}
\end{split} \, .
\label{eq:neglog_laplace_photometric}
\end{equation}
\end{minipage}
\end{adjustbox}

\noindent We integrate the uncertainties in our \ac{projection} formulation as follows:

\begin{equation}
\sigma_{\text{eff}}(p_t) = \sqrt{\Sigma_{t}(p_t)^2 + \Sigma_{t' \rightarrow t}(p_t)^2} \, .
\end{equation}

\noindent This principled combination better identifies regions that violate the static scene assumption in dynamic environments by integrating uncertainty from both frames. We provide details on the proposed \ac{projection} formulation in the supplementary materials.
\subsection{Training Losses}
\label{sec:losses}

\subsubsection{Uncertainty-aware photometric loss} Using our combined projected uncertainty, we define:
\begin{equation}
\mathcal{L}_P = \frac{1}{\left| \mathcal{V} \right|} \sum_{p \in \mathcal{V}} \left( \frac{r\bigl(I_t(p), I_{t' \rightarrow t}(p)\bigr)}
         {\sigma_{\text{eff}}(p)}
+ \log \sigma_{\text{eff}}(p) \right) \, ,
\label{eq: final_photometric_loss_}
\end{equation}
\noindent where $\mathcal{V}$ denotes the set of valid points successfully projected from target to reference image.

\subsubsection{Geometry consistency loss} We adopt the geometry consistency loss from SC-Depth~\cite{bian2021ijcv}, which penalizes inconsistency between projected depth of $p_t$ from target view to reference view, denoted as \( D_{t}^{t'}(p_t) \), and synthesized depth $D_{t' \rightarrow t}(p_t)$:
\begin{equation}
     {D}_{t}^{t'}(p_t) = [T_{t \rightarrow t'} \, D_t(p_t) \, K^{-1} \, p_t]_z \, , \label{eq:depth}
\end{equation}
\begin{equation}
    \mathcal{L}_{\text{Geo}} = \frac{1}{\left| \mathcal{V} \right|} \sum_{p \in \mathcal{V}} \frac{ \left|D_{t' \rightarrow t}(p) - {D}_{t}^{t'}(p) \right| }{ D_{t' \rightarrow t}(p) + {D}_{t}^{t'}(p) } \, , \label{eq:geo}
\end{equation}
where $[\cdot]_z$ denotes the z-coordinate extraction operator.

\subsubsection{Smoothness loss} To address depth discontinuities in low-texture regions, we employ an edge-aware smoothness term on all points on target image and reference image:
\begin{equation}
\mathcal{L}_S = \sum_{p} \left( e^{-\nabla I_t(p)} \cdot \nabla D_t(p) \right)^2 \, .
\end{equation}

\subsubsection{Auto-mask} Following Monodepth2~\cite{monodepth2}, we apply an auto-mask to filter out relative static points moving with the camera:
\begin{equation}
M_a(p) = 
\begin{cases}
1 & \text{if } \|I_t(p) - I_{t' \rightarrow t}(p)\|_1 < \|I_t(p) - I_{t'}(p)\|_1 \\
0 & \text{otherwise}
\end{cases} \, .
\end{equation}

\subsubsection{Training objective} Our final objective combines all losses:
\begin{equation}
\mathcal{L} = M_a \cdot (w_p \mathcal{L}_P + w_g \mathcal{L}_{\text{Geo}}) + w_s \mathcal{L}_S \, ,
\label{eq:final_obejctive}
\end{equation}
\noindent where $w_p$, $w_g$, and $w_s$ are weighting factors for each loss term.


\section{Experiments}

In this section, we evaluate our \ac{method} framework on the KITTI odometry benchmark~\cite{Geiger2012CVPR} and the nuScenes dataset~\cite{nuscenes}. We first present both quantitative and qualitative results for visual odometry, followed by ablation studies to validate the effectiveness of the proposed \ac{projection}. For implementation details and additional results, please refer to the supplementary material.

\subsection{Visual Odometry on KITTI}

\subsubsection{Evaluation metrics} We adopt the relative translation error $ t_{\text{err}} $ (\%), and the relative rotation error $ r_{\text{err}} $ ($^\circ$/100m), following the KITTI odometry benchmark~\cite{Geiger2012CVPR}. We also report the commonly used Absolute Trajectory Error (ATE). Since our method is monocular-based, we perform a 7-DoF alignment to recover a global scale that best aligns the predicted trajectory with the ground-truth for metric computation, following~\cite{bian2021ijcv}.

\subsubsection{Quantitative results}
We compare our \ac{method} against state-of-the-art fully end-to-end unsupervised methods for monocular visual odometry on the KITTI benchmark.

As shown in~\Cref{tab:vo}, Zou et al.~\cite{zou2020learning} achieve the best performance by learning long-term dependencies from video sequences. Our method, \ac{method}, despite using only two consecutive frames, achieves comparable performance to Zou's video-based method and even outperforms it on the ATE metric. Notably, in challenging scenarios like Seq. 01 (highway sequence), where most methods struggle, \ac{method} produces best results, primarily due to our robust \ac{projection} approach for handling dynamic objects. For a quantitative comparison with methods from other \ac{VO} categories, including classical approaches and hybrid approaches (combination of classical and ene-to-end algorithms), please refer to the supplementary materials.

\begin{table}[ht]
    \centering
     \renewcommand{\arraystretch}{1.2} 
    \caption{\textbf{End-to-End Visual Odometry Results on KITTI~\cite{geiger2013vision}.} All methods use monocular training with two consecutive frames, except where $\dagger$ denotes stereo training, and $\ddagger$ indicates video input. "-" signifies unreported values. The best result is shown in \textbf{bold}, and the second-best is \underline{underlined}. We adopt  ATE [m], $t_{\text{err}}$ [\%], and $r_{\text{err}}$ [$^\circ$/100m] as evaluation metrics.}
    \label{tab:vo}
    \resizebox{1\linewidth}{!}{
        \begin{tabular}{l | c c c | c c c | c c c}
            \toprule
            & \multicolumn{3}{c|}{Seq. 01} & \multicolumn{3}{c|}{Seq. 09} & \multicolumn{3}{c}{Seq. 10} \\
            \multirow{-2}{*}{\begin{minipage}[c]{1.5cm}\raggedright Methods\end{minipage}}
            & ATE  & $t_{\text{err}}$  & $r_{\text{err}}$ 
            & ATE  & $t_{\text{err}}$  & $r_{\text{err}}$ 
            & ATE  & $t_{\text{err}}$  & $r_{\text{err}}$ 
            \\
            \midrule
            SfMLearner~\cite{zhou2017unsupervised} & 109.61 & 22.41 & 2.79 & 77.79 & 19.15 & 6.82 & 67.34 & 40.40 & 17.69 \\
            Depth-VO-Feat$\dagger$~\cite{zhan2018unsupervised} & 203.44 & 23.78 & 1.75 & 52.12 & 11.89 & 3.60 & 24.70 & 12.82 & 3.41 \\
            MonoDepth2~\cite{monodepth2} & - & - & - & 76.22 & 17.17 & 3.85 & 20.35 & 11.68 & 5.31 \\
            Zou et al.$\ddagger$~\cite{zou2020learning} & - & - & - & \underline{11.30} & \textbf{3.49} & \textbf{1.00} & \underline{11.80} & \textbf{5.81} & \textbf{1.80} \\
            SC-Depth {\scriptsize (Baseline)}~\cite{bian2021ijcv}& 313.86 & 87.04 & \textbf{1.17} & 26.86 & 7.80 & 3.13 & 13.00 & 7.70 & 4.90 \\
            Manydepth2~\cite{zhou2025manydepth2}& - & - & - & - & 7.01 & \underline{1.76} & - & \underline{7.29} & \underline{2.65} \\
            \midrule
            \textbf{\acs{method} + DINOv2} & \textbf{63.73} & \textbf{19.61} & \underline{1.54} & 14.01 & 4.70 & 1.89 & 13.46 & 7.64 & 3.67 \\
            \textbf{\acs{method} + DepthAnythingV2} & \underline{75.27} & \underline{22.08} & 1.99 & \textbf{9.84} & \underline{4.56} & 2.02 & \textbf{11.28} & 7.76 & 3.58 \\
            \bottomrule
        \end{tabular}
    }
\end{table}

\begin{figure}[!t]
    \centering
    \begin{minipage}[b]{0.43\textwidth}
        \centering
        \includegraphics[width=\textwidth]{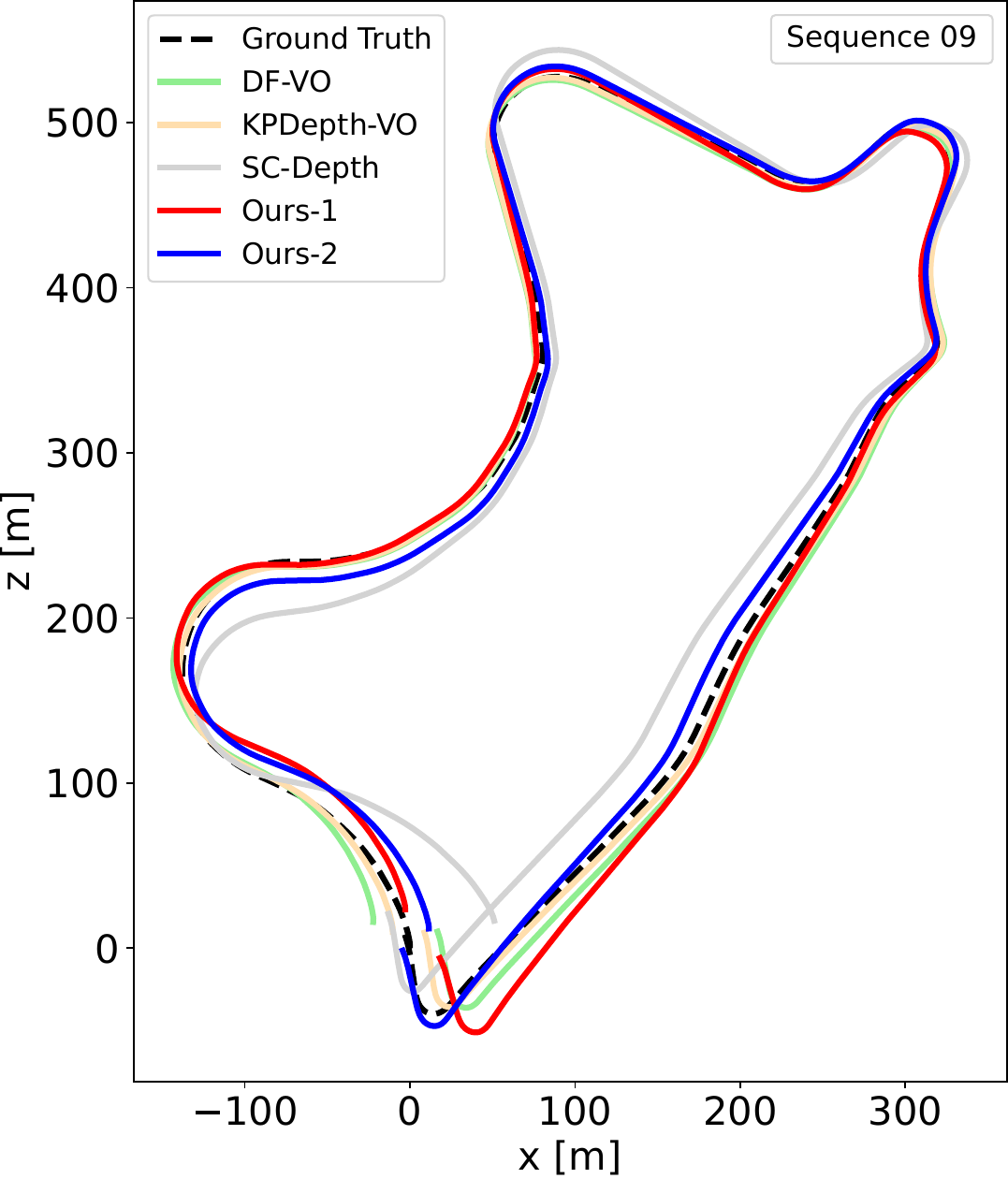}
    \end{minipage}
    \hfill
    \begin{minipage}[b]{0.55\textwidth}
        \centering
        \includegraphics[width=\textwidth]{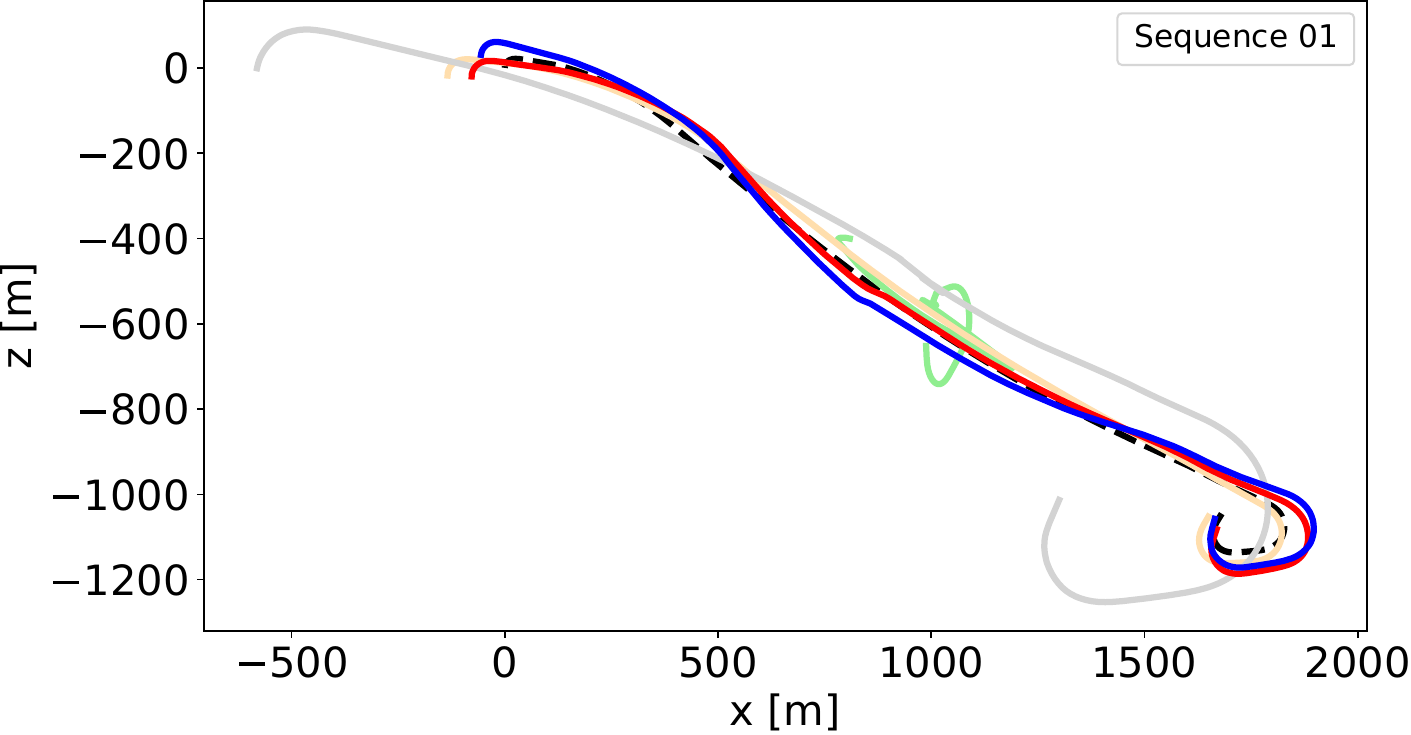}
        \vspace{1em}
        \includegraphics[width=\textwidth]{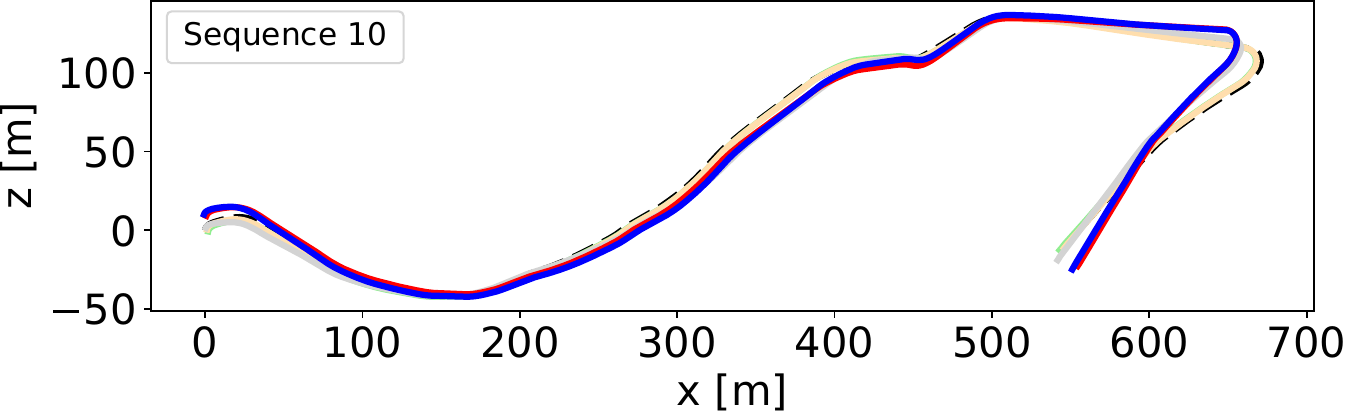}
    \end{minipage}
      \caption{\textbf{Visualization of Trajectories on KITTI~\cite{geiger2013vision}.} In the legend, \textit{Ours-1} represents \ac{method} + DINOv2, and \textit{Ours-2} represents \ac{method} + DepthAnythingV2.}

  \label{fig:viz_kitti}
\end{figure}

\subsubsection{Qualitative results} Visualization of the trajectory results for sequences 01, 09, and 10 are presented in~\Cref{fig:viz_kitti}. In sequence 01, the hybrid method DF-VO~\cite{zhan2020visual} completely fails, while SC-Depth~\cite{bian2021ijcv} exhibits larger deviations compared to our method, which achieves significantly more accurate trajectories. In sequence 09, \ac{method} with DepthAnythingV2~\cite{yang2024depthv2} demonstrates performance comparable to established hybrid methods, including DF-VO~\cite{zhan2020visual} and KP-Depth-VO~\cite{wang2024self}, demonstrating that our approach achieves comparable accuracy to hybrid methods without the computational overhead of their multi-stage processing pipelines.

\subsection{Visual Odometry on nuScenes}

\subsubsection{Evaluation metrics} On the nuScenes dataset~\cite{nuscenes}, we adopt the same ATE metric used in previous experiments on KITTI. However, due to the shorter sequence lengths in nuScenes, we report the frame-to-frame Relative Pose Error (RPE) for translation RPE$_{trans}$ (m) and rotation RPE$_{rot}$ (°) to reflect local accuracy instead of relative translation error.

\begin{figure}[!ht]
\centering
\includegraphics[width=1\textwidth]{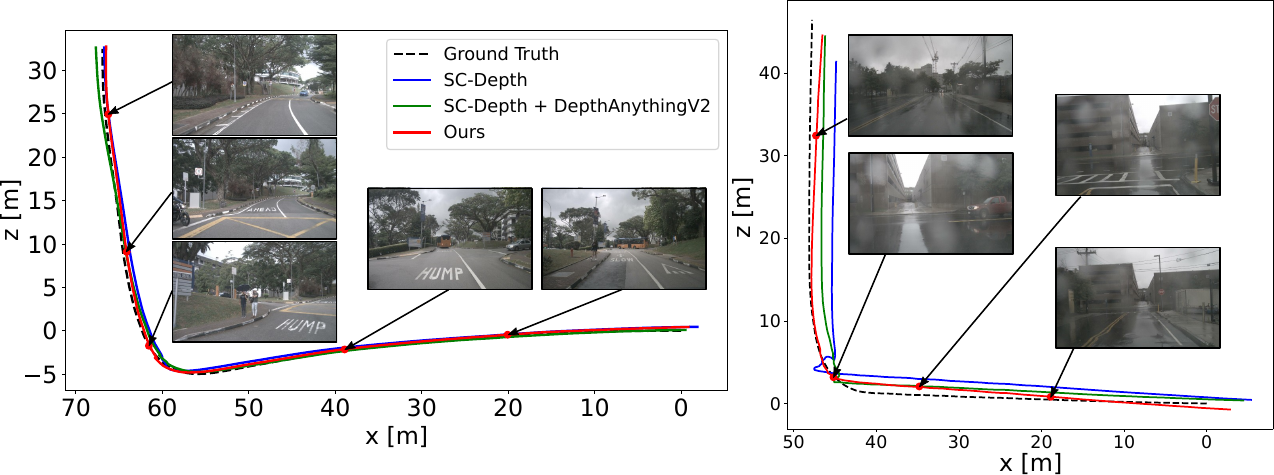}
\caption{\textbf{Visualization of Trajectories on nuScenes~\cite{nuscenes}.} The left and right plots correspond to scene-0928 and scene-0636, respectively. In the legend, \textit{Ours}
represents \acs{method} + DepthAnythingV2. }
\label{fig:trajectories on nuScenes}
\end{figure}

\begin{figure}[!t]
\centering
\includegraphics[width=\textwidth]{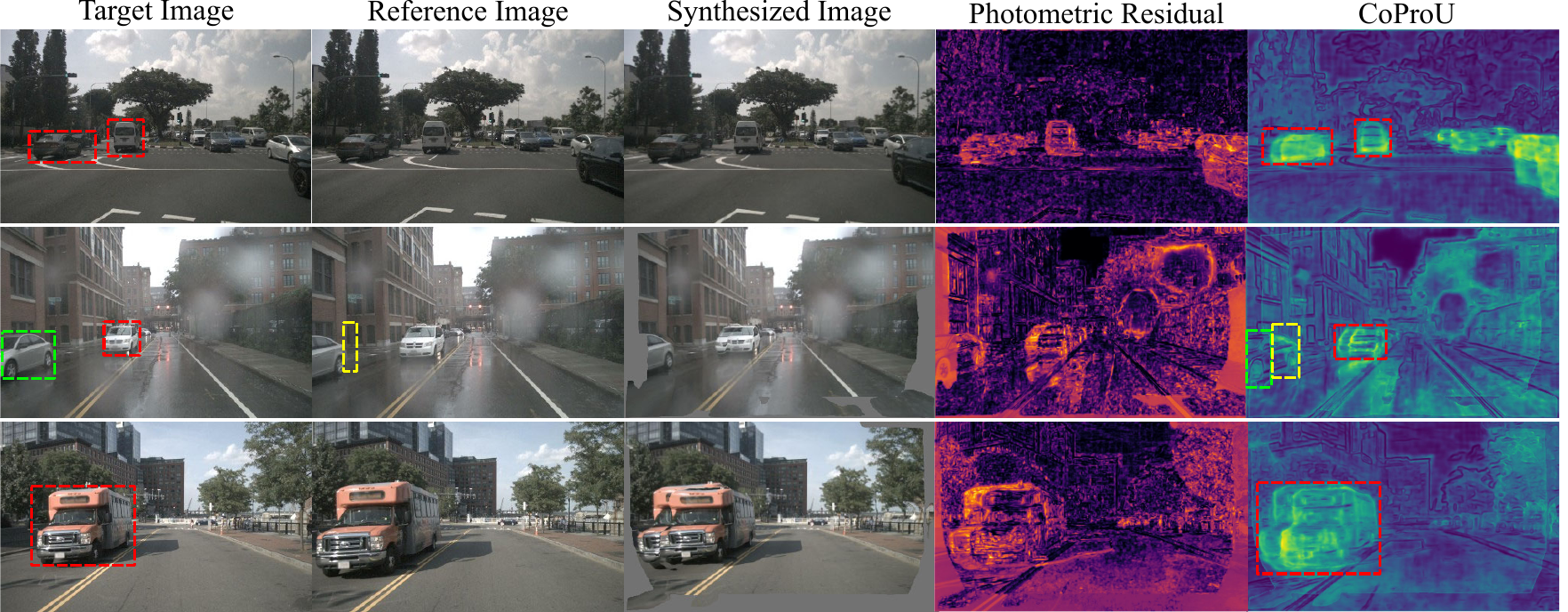}
\caption{\textbf{Uncertainty Visualization on nuScenes~\cite{nuscenes}}. Gray areas in the images indicate invalid regions excluded from loss calculation. Photometric residual brightness represents error magnitude, while \acs{projection} brightness reflects uncertainty. Dynamic objects may appear distorted due to the static scene assumption. Our method robustly masks high-uncertainty regions, distinguishes parked cars (e.g. green box) from moving cars (e.g. red boxes), and detects occluded parts of parked vehicles (e.g. yellow box).}
\label{fig:uncertainty_1}
\end{figure}

\subsubsection{Qualitative results}
As shown in~\Cref{fig:trajectories on nuScenes}, our method predicts more accurate trajectories than the baseline under challenging conditions, such as dynamic scenes and rainy weather. In~\Cref{fig:uncertainty_1}, we observe that the proposed \ac{projection} is able to mask out regions that violate the static scene assumption in dynamic scenes robustly.

\subsubsection{Quantitative results}~\Cref{tab:result_nusc} presents the results on the nuScenes dataset. When the interval between the reference and target images is small, \ac{method} only slightly outperforms the baseline. This is because the nuScenes dataset has a high sampling frequency of 12 Hz, and the vehicle moves at relatively low speeds compared to the KITTI dataset, which leads to limited motion between consecutive frames. To make the scenes more dynamic and better highlight the advantages of our method, we reduce the sampling frequency by increasing the interval between frames. As the interval increases, our method outperforms the baseline by a significantly larger margins.

\begin{table}[!t]
\centering
\caption{\textbf{Visual Odometry Results on nuScenes~\cite{nuscenes}.} We report ATE [m], RPE$_{trans}$ [m], and RPE$_{rot}$ [°]. Our method proves robust and increasingly outperforms the baseline as training and inference intervals increase.}
\label{tab:result_nusc}
\resizebox{1\linewidth}{!}{
\begin{tabular}{l | c | c |c c c | c c c}
\toprule

 & Training & Inference & \multicolumn{3}{c|}{Validation set} & \multicolumn{3}{c}{Test set} \\
\multirow{-2}{*}{\begin{minipage}[c]{1.5cm}\raggedright Methods\end{minipage}} & interval & interval & ATE & RPE$_{trans}$  & RPE$_{rot}$  & ATE & RPE$_{trans}$  & RPE$_{rot}$  \\
\midrule

SC-Depth~\cite{bian2021ijcv}  & 1 & 1 & 0.906 & 0.035 & 0.051 & 0.984 & 0.041 & 0.054  \\
SC-Depth~\cite{bian2021ijcv} + DepthAnythingV2 & 1 & 1 & 0.839 & 0.032 & \textbf{0.048} & 0.910 & 0.039 & 0.052 \\
\acs{projection} + DepthAnythingV2 & 1 & 1 & \textbf{0.826} & \textbf{0.031} & 0.049 & \textbf{0.883} & \textbf{0.037} & \textbf{0.052} \\
\midrule

SC-Depth~\cite{bian2021ijcv}  & 2 & 1 & 0.932 & 0.039 & 0.052 & 1.132 & 0.048 & 0.056 \\
SC-Depth~\cite{bian2021ijcv} + DepthAnythingV2 & 2 & 1 & 0.833 & 0.033 & \textbf{0.046} & 0.877 & 0.037 & 0.051 \\
\acs{projection} + DepthAnythingV2 & 2 & 1 & \textbf{0.771} & \textbf{0.029} & 0.047 & \textbf{0.814} & \textbf{0.033} & \textbf{0.050} \\

\midrule
SC-Depth~\cite{bian2021ijcv}  & 2 & 2 & 0.874 & 0.080 & 0.074 & 1.110 & 0.097 & 0.084 \\
SC-Depth~\cite{bian2021ijcv} + DepthAnythingV2 & 2 & 2 & 0.697 & 0.060 & 0.070 & 0.810 & 0.071 & 0.079 \\
\acs{projection} + DepthAnythingV2 & 2 & 2 & \textbf{0.629} & \textbf{0.054} & \textbf{0.068} & \textbf{0.744} & \textbf{0.065} & \textbf{0.077} \\

\bottomrule
\end{tabular}
}
\end{table}

\subsection{Ablation Study}
\label{sec:ablation_stuty}
\subsubsection{Uncertainty} We validate the advantages of our proposed \ac{projection} in two aspects.
\begin{wrapfigure}[16]{r}{0.6\textwidth}
    \vspace{-2.2em}
    \centering
    \includegraphics[width=\linewidth]{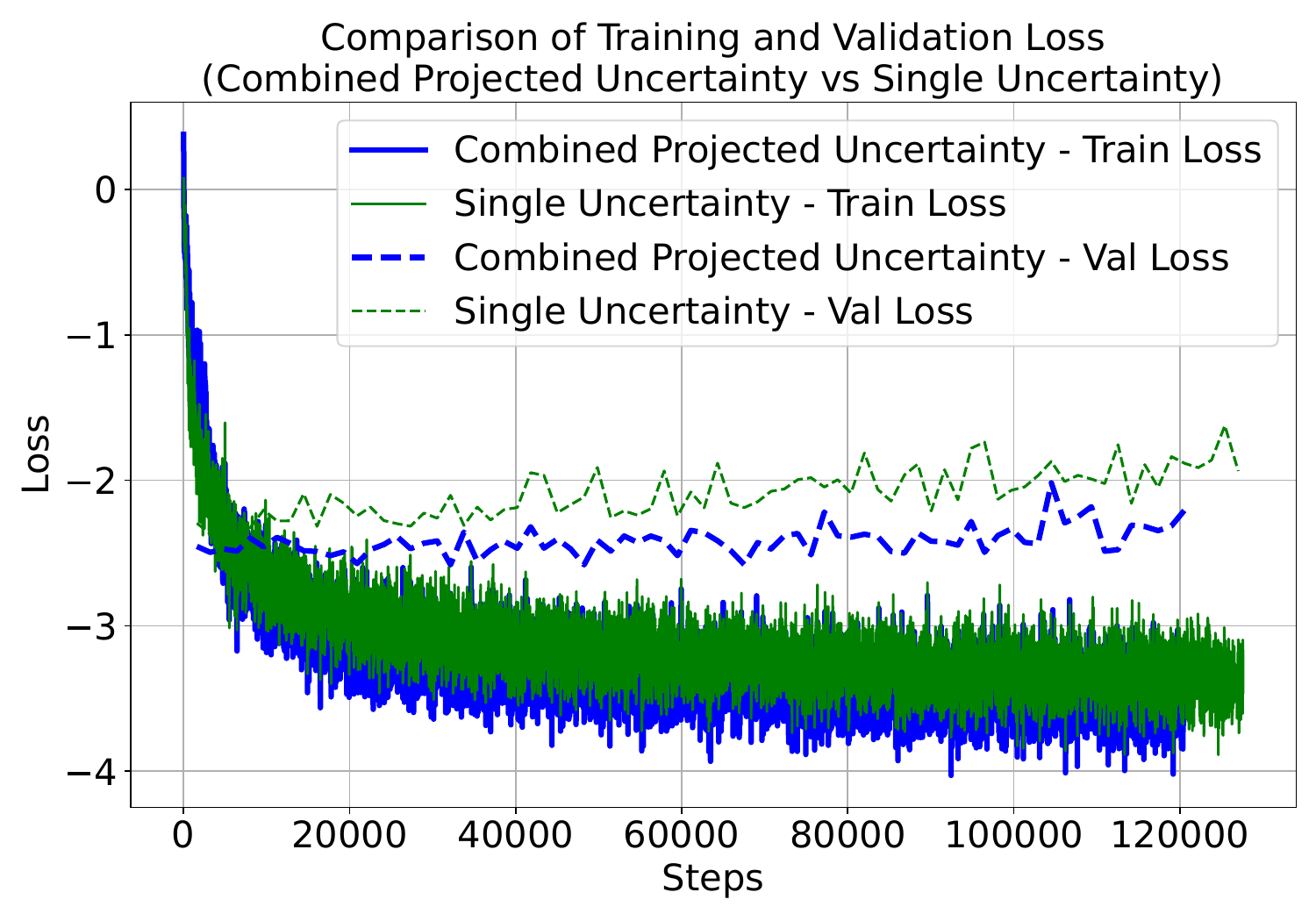}
    \vspace{-2.2em}
      \caption{\textbf{Training and Validation Curves.} Overfitting comparison between \ac{projection} and single uncertainty settings.}
    \label{fig:uncertainty_ablation}
\end{wrapfigure}
First, we show in~\Cref{tab:uncertainty_ablation} that PoseNet trained with \ac{projection} more efficiently masks out dynamic objects, leading to better pose estimation and outperforming PoseNet trained with single uncertainty. 
We verify this by training our model under a single uncertainty setting. For the single uncertainty baseline, following~\cite{yang2020d3vo,wang2024self,wimbauer2025anycam,dai2022self}, we use only the predicted uncertainty $\Sigma_t$ from the target image $I_t$ as the effective uncertainty in~\Cref{eq: final_photometric_loss_}.
Second, as illustrated in ~\Cref{fig:uncertainty_ablation}, training with \ac{projection} results in lower training and validation losses by filtering out more regions that violate the static scene assumption, thereby improving accuracy.

\begin{table*}[t]
\centering
    \caption{\textbf{Ablation Study of the Proposed \acs{projection} on the KITTI Dataset~\cite{geiger2013vision}.} Comparison of \ac{projection} and single uncertainty baseline. We adopt  ATE [m], $t_{\text{err}}$ [\%], and $r_{\text{err}}$ [$^\circ$/100m] as evaluation metrics.}
    \label{tab:uncertainty_ablation}
    \scriptsize
    \renewcommand{\arraystretch}{1.2} 
    \begin{tabular}{l|ccc|ccc}
        \toprule
        & \multicolumn{3}{c|}{Seq. 09} & \multicolumn{3}{c}{Seq. 10} \\ 
        \multirow{-2}{*}{\begin{minipage}[c]{1.5cm}\raggedright Methods\end{minipage}} & ATE  & $t_{\text{err}}$  & $r_{\text{err}}$  
        & ATE & $t_{\text{err}}$  & $r_{\text{err}}$ \\ 
        \midrule
        Single uncertainty + DepthAnythingV2  & 17.06 & 8.35 & 3.16 & 15.62 & 10.54 & 4.05 \\ 
        \acs{projection} + DepthAnythingV2 & \textbf{9.84} & \textbf{4.56} & \textbf{2.02} & \textbf{11.28} & \textbf{7.76} & \textbf{3.58} \\ 
        \bottomrule
    \end{tabular}
\end{table*}

\subsubsection{DepthNet backbone} Unlike previous end-to-end works~\cite{monodepth2,zhou2017unsupervised} and our baseline SC-Depth~\cite{bian2021ijcv} that use ResNet~\cite{he2016deep} for depth prediction, we employ the Vision Transformer Small (ViT-S) from either DINOv2~\cite{oquab2024dinov2} or DepthAnythingV2~\cite{yang2024depthv2}. To isolate the effect of our \ac{projection} from backbone improvements, we conduct an ablation study with results presented in~\Cref{tab:backbone_ablation}.

\begin{table}[t]
\centering
\renewcommand{\arraystretch}{1} 
\caption{\textbf{Ablation Study on the Effect of the DepthNet Backbone on the KITTI Dataset~\cite{geiger2013vision}.} The ViT-S is from DepthAnythingV2~\cite{yang2024depthv2}. The self-discovered mask refers to the dynamic mask proposed in SC-Depth~\cite{bian2021ijcv}. We report ATE [m], $t_{\text{err}}$ [\%], and $r_{\text{err}}$ [$^\circ$/100m].}

\label{tab:backbone_ablation}
\resizebox{1\linewidth}{!}{
\begin{tabular}{l|c|c|c|ccc|ccc}

\toprule
  & & & How to mask & \multicolumn{3}{c|}{Seq. 09} & \multicolumn{3}{c}{Seq. 10} \\
\multirow{-2}{*}{\begin{minipage}[c]{1.5cm}\raggedright Methods\end{minipage}} & \multirow{-2}{*}{\begin{minipage}[c]{1.5cm}\centering Backbone\end{minipage}}  & \multirow{-2}{*}{\begin{minipage}[c]{1.5cm}\centering Frozen\end{minipage}} & dynamic objects? & ATE  & $t_{\text{err}}$  & $r_{\text{err}}$  & ATE  & $t_{\text{err}}$  & $r_{\text{err}}$  \\
\midrule
SC-Depth~\cite{bian2021ijcv} & ResNet-50 & no & self-discovered mask & 26.86 & 7.80 & 3.13 & 13.00 & 7.70 & 4.90 \\
SC-Depth~\cite{bian2021ijcv} & ViT-S & yes & self-discovered mask & 10.75 & 5.47 & \textbf{1.87} & 13.74 & 8.84 & \textbf{3.44} \\
\acs{method} (ours) & ViT-S & yes & \acs{projection} & \textbf{9.84} & \textbf{4.56} & 2.02 & \textbf{11.28} & \textbf{7.76} & 3.58 \\
\bottomrule
\end{tabular}
}
\end{table}

When replacing ResNet-50 in SC-Depth~\cite{bian2021ijcv} with ViT-S from DepthAnythingV2~\cite{yang2024depthv2}, performance on Seq. 09 improves significantly. Applying \ac{projection} leads to further improvements, confirming that our uncertainty approach contributes substantially to the overall performance gains. A similar performance improvement is observed in~\Cref{tab:result_nusc} for the nuScenes~\cite{nuscenes} dataset.

\section{Conclusion}

In this paper, we presented \ac{method}, a robust visual odometry system introducing a novel combined projected uncertainty approach. Our method effectively masks out regions that violate the static scene assumption from both target and reference images, significantly outperforming existing two-frame-based end-to-end \ac{VO} methods on the KITTI odometry benchmark and nuScenes dataset, particularly in challenging scenarios like highway driving and dynamic scenes.

\subsubsection{Limitations and future work} While effective, \ac{method} has limitations. The ResNet-18 backbone in PoseNet restricts representation capacity, suggesting the need for more advanced architectures. Frame-independent uncertainty prediction would benefit from better inter-frame modeling, especially in dynamic scenes. While our method shows significant improvement in relative translation error, the gains in absolute trajectory error are more modest. Future work will address these challenges, utilize larger datasets, refine our occlusion handling approach, and employ unfrozen backbones to further enhance monocular visual odometry.

\ifbool{true}{
\appendix
\section*{Appendix}

\renewcommand{\theequation}{A\arabic{equation}}

\renewcommand{\thefigure}{A\arabic{figure}}

\renewcommand{\thetable}{A\arabic{table}}

\section{Overview}
In this supplementary material, we first explore potential future directions in \Cref{app:future_work}. Next, we address the limitations of our methods in \Cref{app:limitations} and ethical and safety considerations in \Cref{app:ethical_concerns}. Following this, we present the \ac{projection} formulation in \Cref{app:formulation}. Detailed implementation descriptions are then provided in \Cref{app:imp_details} to facilitate the reproduction of our experimental results, and additional results and visualizations are presented in \Cref{app:results}.

\section{Future Work}
\label{app:future_work}
In future work, we aim to address several key areas to further enhance the performance of our method. First, we plan to explore more powerful architectures for pose estimation to overcome the limitations of the ResNet-18 backbone in our PoseNet, which partly contributes to deviations in trajectory estimation. Second, we intend to develop uncertainty prediction approaches that incorporate information from multiple frames simultaneously, enabling better inter-frame relationship modeling and improving the handling of dynamic objects in complex scenes. Finally, \ac{method} can be integrated with geometry reconstruction methods such as DUSt3R~\cite{wang2024dust3r} and VGGT~\cite{wang2025vggt} to train them in an unsupervised fashion. 

\section{Limitations}
\label{app:limitations}

Despite achieving good results and notable improvements in several metrics for monocular unsupervised end-to-end \ac{VO} methods, our approach has certain limitations that warrant further exploration. First, compared to hybrid methods with non-differentiable optimization back-ends, our method exhibits some deviations in trajectory estimation, partly due to the representation capacity of the ResNet-18 backbone in our PoseNet. 
Second, our current uncertainty prediction mechanism operates independently for each image, which limits its ability to effectively model inter-frame relationships. This affects its performance, particularly in handling dynamic objects in complex scenes. 
Third, our method struggles when photometric residuals are low, causing dynamic elements to receive low uncertainties. Objects with similar ego-motions as shown in \Cref{fig:infinity_depth} or those maintaining appearance consistency across frames despite motion are assigned low uncertainties as shown in \Cref{fig:low_photometric_error}. 
Fourth, our method provides trajectory estimations only up to a scale, which limits its applicability in scenarios where absolute scale recovery is critical. Finally, while our method shows significant improvement in translation error, we observe larger rotation errors compared to the baseline. This occurs because the model tends to assign higher uncertainty to near-range pixels, which sometimes leads to the overlooking of valuable geometric information crucial for accurate rotation estimation.

\begin{figure}[!t]
    \centering
    \includegraphics[width=1\textwidth]{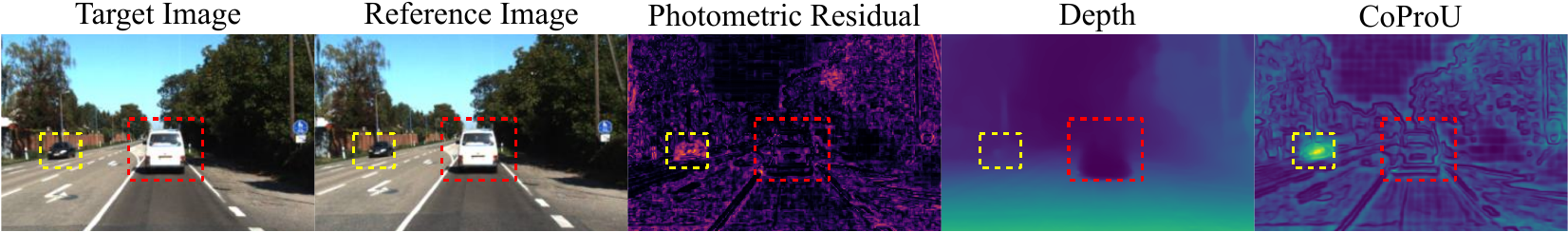}
    \vspace{-1\baselineskip}
   \caption{\textbf{Depth Estimation Failure for Similar Ego-Motion Object.} The object (red box) moving along the same trajectory as the camera exhibits minimal photometric changes, causing the depth estimation to treat it as a static object similar to distant elements like the sky. Consequently, it is assigned extremely large depth values and receives low uncertainty estimates that fail to identify it as dynamic despite its actual motion. In contrast, another dynamic object (yellow box) at a similar distance was successfully assigned high uncertainty due to its high photometric residual, yielding accurate depth prediction and correct uncertainty estimation.}
    \label{fig:infinity_depth}
\end{figure}

\begin{figure}[!t]
    \centering
    \includegraphics[width=1\textwidth]{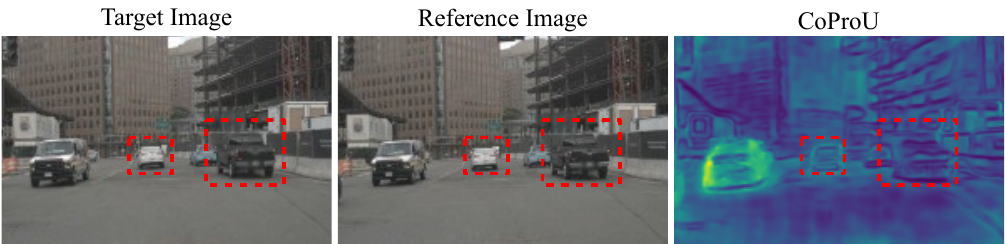}
    \vspace{-1\baselineskip}
    \caption{\textbf{Failure to Detect Moving Objects with Low Photometric Error.} The moving objects (red boxes) exhibit low photometric residuals, leading to low uncertainty estimates that fail to identify them as dynamic.}
    \label{fig:low_photometric_error}
\end{figure}

\section{Ethical and Safety Considerations}
\label{app:ethical_concerns}
Our study relies on open-source datasets commonly used in research. These datasets are provided with the responsibility of adhering to privacy-preserving practices, including the anonymization of sensitive information such as faces and license plates. 
Also, it should be noted that there is no guarantee that our method generalizes effectively to unknown environments where the traffic conditions differ significantly from those present in the datasets used for training. Additionally, our method does not achieve the centimeter-level accuracy required by most safety-critical applications. As a result, it is currently not adequate for deployment in high-precision safety applications where reliability is paramount.

\section{\ac{projection} Formulation}
\label{app:formulation}
In this section, we present the mathematical foundation of our proposed \ac{projection}. We begin by reviewing how supervised methods model heteroscedastic aleatoric uncertainty. We then explain why modeling uncertainty with a single source fails from a mathematical perspective, and finally describe how our method addresses this limitation.

\subsubsection{Supervised scenario} In supervised methods, given the ground-truth $y$ and the prediction $\hat{y}$, the heteroscedastic aleatoric uncertainty in the data, denoted by $\sigma$, is modeled using a Laplace distribution~\cite{ilg2018uncertainty,yang2020d3vo,poggi2020uncertainty,kendall2017uncertainties}:
\begin{equation}
p(y \mid \hat{y}, \sigma) = \frac{1}{2\sigma} \exp\left(-\frac{|y - \hat{y}|}{\sigma}\right) \, .
\end{equation}
\noindent Taking the negative logarithm of the likelihood gives:
\begin{equation}
- \log p(y \mid \hat{y}, \sigma) = \frac{|y - \hat{y}|}{\sigma} + \log \sigma + \text{const} \, .
\end{equation}
\noindent Notably, the formulated uncertainty is associated solely with the ground-truth $y$, as discussed in~\cite{kendall2017uncertainties}; the prediction $\hat{y}$ is assumed to be deterministic and carries no uncertainty.

\subsubsection{Single uncertainty formulation failing} D3VO~\cite{yang2020d3vo} adopts the above formulation into unsupervised \ac{VO} by treating the target image $I_t$ as the ground-truth with heteroscedastic aleatoric uncertainty, while considering the synthesized image $I_{t' \rightarrow t}$ as a deterministic prediction without any uncertainty. However, this naive adoption neglects the fact that the synthesized image, which is derived from the reference image $I_{t'}$, also inherently carries heteroscedastic aleatoric uncertainty.

\subsubsection{Problem definition} In contrast to the single uncertainty formulation, we aim to account for uncertainty arising from both the target and reference images. To this end, we model each pixel intensity as an independent Laplace random variable: $I_t(p_t) \sim \mathrm{Laplace}\bigl(\mu_1,\Sigma_t(p_t)\bigr)$ and $I_{t'\rightarrow t}(p_t) \sim \mathrm{Laplace}\bigl(\mu_2,\Sigma_{t'\rightarrow t}(p_t)\bigr)$, where $\mu_1$ and $\mu_2$ represent the underlying true values. Our objective is to model the likelihood of the photometric residual $r\bigl(I_t(p_t), I_{t' \rightarrow t}(p_t)\bigr)$. However, obtaining a closed-form expression is challenging due to three factors: (1) the residual computation includes \ac{SSIM}~\cite{wang2004image} components; (2) the $L_1$ difference between two independent Laplace-distributed variables does not itself follow a Laplace distribution; and (3) the true underlying values $\mu_1$ and $\mu_2$ are unknown. As a result, we approximate the residual likelihood using a Laplace distribution with effective uncertainty to obtain the simple supervision signal defined in~\Cref{eq:laplace_photometric}.

\subsubsection{Effective uncertainty design} We define the effective uncertainty $\sigma_{\text{eff}}(p_t)$ at point $p_t$ in~\Cref{eq:laplace_photometric} based on the variance of the joint distribution resulting from the difference between two independent Laplace-distributed variables. Let

\begin{adjustbox}{width=1\linewidth}
\begin{minipage}{1.1\linewidth}
\begin{equation}
A \sim \operatorname{Laplace}(\mu_{1}, \sigma_{1}), \;
B \sim \operatorname{Laplace}(\mu_{2}, \sigma_{2}), \quad 
\text{with } 
A \perp B, \;
\sigma_{1}, \sigma_{2} > 0, \; \mu_{1} \neq \mu_{2} \, .
\end{equation}
\end{minipage}
\end{adjustbox}

\paragraph{1. Individual Densities.}
The probability density functions are given by:
\[
f_{A}(a) = \frac{1}{2\sigma_1} e^{- \frac{|a - \mu_1|}{\sigma_1}}, \qquad
f_{B}(b) = \frac{1}{2\sigma_2} e^{- \frac{|b - \mu_2|}{\sigma_2}} \,.
\]

\paragraph{2. Convolution for the Residual \( R \).}
Since \( R = A - B \), the density function of \( R \) is given by the convolution:
\begin{equation}
\label{eq: joint_convolution}
f_R(r) = \int_{-\infty}^{\infty} f_A(a) f_B(a - r) \, da \,.
\end{equation}

\paragraph{3. Change of Variables.}
Let
\[
u = a - \mu_1, \qquad
\Delta = \mu_1 - \mu_2, \qquad
c = r - \Delta \,,
\]
so that \( a = \mu_1 + u \), and \( a - r - \mu_2 = u - c \). Substituting into~\Cref{eq: joint_convolution} gives:
\begin{equation}
f_R(r) = \frac{1}{4\sigma_1 \sigma_2} \int_{-\infty}^{\infty} 
e^{- \frac{|u|}{\sigma_1} - \frac{|u - c|}{\sigma_2}} \, du \,.
\end{equation}

\paragraph{4. Probability Density Function.}
By evaluating the piecewise integral (omitted here), we obtain:
\begin{equation}
f_R(r) =
\frac{\sigma_2 \, e^{- \frac{|r - \Delta|}{\sigma_2}} 
- \sigma_1 \, e^{- \frac{|r - \Delta|}{\sigma_1}}}
{2(\sigma_2^2 - \sigma_1^2)}, \qquad \text{for } \sigma_1 \neq \sigma_2 \,.
\end{equation}
Note that this is not a Laplace distribution.

\paragraph{5. Expectation.}
\begin{equation}
\mathbb{E}[R] = \int_{-\infty}^{\infty} r f_R(r) \, dr = \mu_1 - \mu_2 \, .
\end{equation}

\paragraph{6. Variance.}
\begin{equation}
\mathrm{Var}(R) = \int_{-\infty}^{\infty} (r - \mathbb{E}[R])^2 f_R(r) \, dr = 2(\sigma_1^2 + \sigma_2^2) \, .
\end{equation}

\paragraph{7. Laplace Approximation.} From Steps 5 and 6, we have:
\[
\mathbb{E}[R] = \mu_1 - \mu_2, \qquad \mathrm{Var}(R) = 2(\sigma_1^2 + \sigma_2^2) \, .
\]
Since the variance of a Laplace distribution $\mathrm{Laplace}(\mu, \sigma)$ is given by $2\sigma^2$, we equate:
\[
2\sigma_{\text{eff}}^2 = 2(\sigma_1^2 + \sigma_2^2) \quad \Rightarrow \quad \sigma_{\text{eff}} = \sqrt{\sigma_1^2 + \sigma_2^2} \, .
\]
Thus, the residual distribution can be approximated by:
\begin{equation}
f_R(r) \approx \frac{1}{2\sigma_{\text{eff}}} e^{- \frac{|r - (\mu_1 - \mu_2)|}{\sigma_{\text{eff}}}}, \qquad
\sigma_{\text{eff}} = \sqrt{\sigma_1^2 + \sigma_2^2} \, .
\end{equation}
\paragraph{8. Pixel-wise Substitution.}
To apply this approximation in our pixel-wise formulation, we substitute the heteroscedastic aleatoric uncertainties from the target and synthesized images:
\begin{equation}
\sigma_{\text{eff}}(p_t) = \sqrt{\Sigma_{t}(p_t)^2 + \Sigma_{t' \rightarrow t}(p_t)^2} \, .
\end{equation}
Nearing the optimum, the expected residual at each pixel should approach zero, i.e., \( \mathbb{E}[r\bigl(I_t(p_t), I_{t' \rightarrow t}(p_t)\bigr)] = 0 \).
Therefore, we can safely assume that the mean of the residual distribution is zero, which further simplifies the likelihood approximation. Under this assumption and by substituting the uncertainty terms into the Laplace likelihood, we finally approximate the photometric residual likelihood at each pixel as in~\Cref{eq:laplace_photometric}.

\section{Implementation Details}
\label{app:imp_details}
\subsubsection{Network architecture} For the pre-trained Vision Transformer~\cite{dosovitskiy2020image}, we use either DINOv2~\cite{oquab2024dinov2} ViT-Small or DepthAnythingV2 ViT-Small~\cite{yang2024depthv2}. For the DPT layer~\cite{ranftl2021vision}, we follow the modified version introduced in DepthAnythingV2~\cite{yang2024depthv2}. Furthermore, we replace the last activation function from ReLU to Sigmoid and change the number of final output channels from one to two, corresponding to disparity and uncertainty. The disparity $ x $ is then converted to depth via $ D = 1 / (a x + b) $, where $ a $ and $ b $ are selected to constrain $ D $ within the range of $[0.1, 100]$ units, which is a common setting for outdoor driving scenarios. Following~\cite{bian2021ijcv}, for the pose network, we use ResNet-18~\cite{he2016deep}, where the first layer is modified to accept six-channel inputs by concatenating two RGB images. The encoded features from ResNet-18 are subsequently decoded using four convolutional layers into 6-DoF pose parameters.

\subsubsection{Dataset} We use the KITTI odometry dataset~\cite{geiger2013vision,Geiger2012CVPR} and nuScenes dataset~\cite{nuscenes}. For KITTI, Sequences 00–08 are used for training, with sequence 01 reserved for validation. During training, we utilize videos from both the left and right cameras, treating them as two independent monocular videos rather than as a stereo pair, resulting in 38K training image pairs. All images are resized to a resolution of $ 256 \times 832 $. For testing, we use sequences 09–10.

For the nuScenes dataset~\cite{nuscenes}, we exclude all scenes recorded at night. Since nuScenes does not provide a VO benchmark on test set, we construct one by splitting the original validation set into separate validation and test subsets. This results in 616 scenes for training and 135 scenes for validation and testing. Among the 135 scenes, we allocate the first 70 for validation and the remaining 65 for testing. All images are resized to a resolution of $ 256 \times 416 $.

\subsubsection{Two-frames input}
For experiments on KITTI~\cite{geiger2013vision}, we extract consecutive image pairs $(I_t, I_{t-1})$ and $(I_t, I_{t+1})$ from the original videos. For the nuScenes dataset~\cite{nuscenes}, we additionally experiment with larger temporal intervals, extracting pairs $(I_t, I_{t-s})$ and $(I_t, I_{t+s})$ where $s \geq 1$. This extended temporal spacing allows us to evaluate model performance under conditions that simulate lower frame rates or scenarios with significant ego motion between frames.

\subsubsection{Training losses}
We use \ac{SSIM}~\cite{wang2004image}, defined between two patches $x$ and $y$ as:
\begin{equation}
\operatorname{SSIM}(x,y)=
\frac{\bigl(2\mu_x\mu_y + C_1\bigr)\bigl(2\sigma_{xy}+C_2\bigr)}
     {\bigl(\mu_x^{2}+\mu_y^{2}+C_1\bigr)\bigl(\sigma_x^{2}+\sigma_y^{2}+C_2\bigr)} \, ,
\label{eq:ssim}
\end{equation}
\noindent where $\mu$ and $\sigma$ are local means and standard deviations, and $C_1, C_2$ are small stabilizing constants.

\subsubsection{Training setting} For the KITTI dataset~\cite{geiger2013vision}, we set $C_1 = 0.0009$, $C_2 = 0.0001$, and $\alpha = 0.85$ in~\Cref{eq:photometric,eq:ssim}. In the total loss equation, we use $w_p = 1$, $w_g = 0.5$, and $w_s = 0.1$. During training, images are augmented with random scaling, cropping, and horizontal flipping. We use the AdamW optimizer~\cite{loshchilov2017decoupled} to train the model for 90 epochs with a batch size of 24. The learning rate is set to $5 \times 10^{-4}$ for the first 75 epochs and decreased to $2.5 \times 10^{-4}$ for the final 15 epochs. The model was trained on the KITTI odometry dataset using one NVIDIA A40 GPU for approximately 2.5 days.

For the nuScenes dataset~\cite{nuscenes}, we train \ac{method} for 20 epochs using a learning rate of $5 \times 10^{-4}$ and a batch size of 36. All other hyperparameters follow the training settings used for KITTI~\cite{geiger2013vision}. Additionally, we train SC-Depth~\cite{bian2021ijcv} using either ResNet-50~\cite{he2016deep} or DepthAnythingV2~\cite{yang2024depthv2} as the backbone on nuScenes. For these models, we set the learning rate to $1 \times 10^{-4}$, as higher values (e.g., $5 \times 10^{-4}$) lead to training divergence, verifying the capacity of \ac{method} to enable more stable training. Notably, a learning rate of $1 \times 10^{-4}$ is also the default configuration in SC-Depth~\cite{bian2021ijcv}. We trained all models on the nuScenes dataset for one day each using a single NVIDIA A40 GPU.

\subsubsection{Inference} During inference, we only need to predict the relative pose between consecutive images. Since the PoseNet is independent in our framework, we simply pass the input images through the Lightweight PoseNet, enabling real-time performance.

\section{Additional Results}
\label{app:results}
\subsubsection{Comparison with Additional Types of Unsupervised VO} 
Existing unsupervised monocular approaches fall into three categories: classical methods using traditional geometric algorithms, end-to-end methods using only neural networks, and hybrid methods combining classical geometry with learning-based components.  Based on training data, methods are classified as monocular or stereo, with stereo-trained methods not necessarily requiring stereo input during inference. Methods are further categorized as video-based or two-frame-based depending on input sequence length. ~\Cref{tab:vo_additional} presents the full comparison between \ac{method} and approaches in all unsupervised categories.

\begin{table}[t]
\centering
\renewcommand{\arraystretch}{1.2}
\caption{\textbf{Visual Odometry Results on KITTI~\cite{geiger2013vision}.} "Types" indicates the model design (Classical / Hybrid / End-to-End). "Input Type" denotes V (video) or TF (two consecutive frames). "Training Camera" specifies M (monocular), S (stereo), or vS (virtual stereo). In both the hybrid and end-to-end categories, the best results are highlighted in \textbf{bold}, while the second-best results are \underline{underlined}. We present the ATE~[m], $t_{\text{err}}$~[\%], and $r_{\text{err}}$~[$^\circ$/100m].}
\label{tab:vo_additional}
\resizebox{\linewidth}{!}{
    \begin{tabular}{c l | c c | c c c | c c c | c c c}
        \toprule
        \multirow{2}{*}{Types} & \multirow{2}{*}{Methods} & \multirow{2}{*}{Input} & \multirow{2}{*}{Training} & \multicolumn{3}{c|}{Seq. 01} & \multicolumn{3}{c|}{Seq. 09} & \multicolumn{3}{c}{Seq. 10} \\
        & & & & ATE & $t_{\text{err}}$ & $r_{\text{err}}$ & ATE & $t_{\text{err}}$ & $r_{\text{err}}$ & ATE & $t_{\text{err}}$ & $r_{\text{err}}$ \\
        \midrule
        \multirow{2}{*}{Classical} & ORB-SLAM2 (w/o loop closure)~\cite{mur2017orb} & V & - & 502.20 & 107.57 & 0.89 & 38.77 & 9.30 & 0.26 & 5.42 & 2.57 & 0.32 \\
        & ORB-SLAM2 (w loop closure)~\cite{mur2017orb} & V & - & 508.34 & 109.10 & 0.45 & 8.39 & 2.88 & 0.25 & 6.63 & 3.30 & 0.30 \\
        \midrule
        \multirow{8}{*}{Hybrid} & DVSO~\cite{yang2018deep} & V & S & - & - & - & - & \underline{0.83} & \textbf{0.21} & - & \underline{0.74} & \textbf{0.21} \\
        & Dai et al.~\cite{dai2022self} & V & M & - & - & - & - & 3.24 & 0.87 & - & 1.03 & 0.65 \\
        & D3VO~\cite{yang2020d3vo} & V & S & - & - & - & \textbf{2.68} & \textbf{0.78} & - & \textbf{0.87} & \textbf{0.62} & - \\
        & VRVO~\cite{zhang2022towards} & V & vS & - & - & - & \underline{4.39} & 1.55 & 0.28 & 6.04 & 2.75 & 0.36 \\
        & DFVO~\cite{zhan2020visual} & TF & M & 117.40 & 39.46 & \textbf{0.50} & 8.36 & 2.40 & \underline{0.24} & 3.13 & 1.82 & 0.38 \\
        & DualRefine-refined~\cite{bangunharcana2023dualrefine} & TF & M & - & - & - & 5.18 & 3.43 & 1.04 & 10.85 & 6.80 & 1.13 \\
        & KPDepth-VO~\cite{wang2024self} & TF & M & \textbf{54.08} & \textbf{15.03} & 0.52 & 6.06 & 1.86 & 0.29 & \underline{2.64} & 1.55 & \underline{0.31} \\
        \midrule
        \multirow{9}{*}{\begin{tabular}{c}End\\to\\End\end{tabular}} & SfMLearner~\cite{zhou2017unsupervised} & TF & M & 109.61 & 22.41 & 2.79 & 77.79 & 19.15 & 6.82 & 67.34 & 40.40 & 17.69 \\
        & Depth-VO-Feat~\cite{zhan2018unsupervised} & TF & S & 203.44 & 23.78 & 1.75 & 52.12 & 11.89 & 3.60 & 24.70 & 12.82 & 3.41 \\
        & GeoNet~\cite{yin2018geonet} & TF & M & - & - & - & 158.45 & 28.72 & 9.80 & 43.04 & 23.90 & 9.00 \\
        & MonoDepth2~\cite{monodepth2} & TF & M & - & - & - & 76.22 & 17.17 & 3.85 & 20.35 & 11.68 & 5.31 \\
        & Zou et al.~\cite{zou2020learning} & V & M & - & - & - & \underline{11.30} & \textbf{3.49} & \textbf{1.00} & \underline{11.80} & \textbf{5.81} & \textbf{1.80} \\
        & SC-Depth~\cite{bian2021ijcv} & TF & M & 313.86 & 87.04 & \textbf{1.17} & 26.86 & 7.80 & 3.13 & 13.00 & 7.70 & 4.90 \\
        & Manydepth2~\cite{zhou2025manydepth2} & TF & M & - & - & - & - & 7.01 & \underline{1.76} & - & \underline{7.29} & \underline{2.65} \\
        & \textbf{\acs{method} + DINOv2} & TF & M & \textbf{63.73} & \textbf{19.61} & \underline{1.54} & 14.01 & 4.70 & 1.89 & 13.46 & 7.64 & 3.67 \\
        & \textbf{\acs{method} + DepthAnythingV2} & TF & M & \underline{75.27} & \underline{22.08} & 1.99 & \textbf{9.84} & \underline{4.56} & 2.02 & \textbf{11.28} & 7.76 & 3.58 \\
        \bottomrule
    \end{tabular}
}
\end{table}

The table shows that hybrid methods generally outperform both classical and fully end-to-end approaches, benefiting from geometry-based refinement or optimization back-ends. For example, D3VO~\cite{yang2020d3vo}, a state-of-the-art self-supervised monocular VO method, leverages stereo training for real-world scale and uncertainty estimation, and refines predictions through bundle adjustment~\cite{engel2017direct} for accurate camera tracking.

Despite their strong performance, our work focuses on fully end-to-end methods for three reasons: (1) hybrid approaches usually rely on feature matching and geometry optimization, which are computationally intensive and limit real-time applicability; (2) these steps increase training complexity and reduce scalability on large datasets; and (3) recent successes such as VGGT~\cite{wang2025vggt} and DUSt3R~\cite{wang2024dust3r} suggest that end-to-end methods can potentially surpass hybrid ones with large-scale data.

Notably, both D3VO~\cite{yang2020d3vo} and KPDepth-VO~\cite{wang2024self} adopt single-uncertainty mechanisms to mask dynamic regions. Our proposed \ac{projection} could be integrated into these methods to boost their performance. As shown in~\Cref{sec:ablation_stuty} and in this supplementary, our \ac{projection} outperforms single uncertainty, indicating its potential to further improve these hybrid frameworks.

\subsubsection{Inference speed}
We verify the real-time capability of \ac{method} by measuring frames per second (FPS) during inference with PoseNet. The results are presented in~\Cref{tab:inference_speed_ours}. Under identical hardware conditions, our method achieves a substantial \SI{20}{\times} speedup compared to the state-of-the-art hybrid monocular method KPDepth-VO~\cite{wang2024self}, which reaches only 12.86~FPS due to computational overhead from GPU-to-CPU data transfers and subsequent CPU-based optimization steps.

\begin{table}[t]
\centering
\scriptsize
\renewcommand{\arraystretch}{1} 
\caption{\textbf{Pose Inference Speed of \acs{method}.} The experiments were conducted using an NVIDIA RTX A2000 GPU and an Intel Xeon E3-1226 v3 CPU (3.30GHz, 4 cores).}
\label{tab:inference_speed_ours}

\begin{tabular}{c|c|c|c|c}
\toprule
Dataset & Backbone & Resolution & Device & FPS \\
\midrule
\multirow{2}{*}{KITTI~\cite{geiger2013vision}} 
& \multirow{2}{*}{ResNet-18} 
& \multirow{2}{*}{256$\times$832} 
& GPU & 247.88 \\
& & & CPU & 11.46 \\
\midrule
\multirow{2}{*}{nuScenes~\cite{nuscenes}} 
& \multirow{2}{*}{ResNet-18} 
& \multirow{2}{*}{256$\times$416} 
& GPU & 352.90 \\
& & & CPU & 20.48 \\
\bottomrule
\end{tabular}
\end{table}

\subsubsection{Ablation study on uncertainty prediction in nuScenes} Table~\ref{tab:uncertainty_ablation_nusc} shows that our projection method outperforms the single uncertainty baseline on the nuScenes~\cite{nuscenes} dataset. Additionally, our method remains robust across different inference intervals and shows improvements as intervals increase.

\begin{table}[ht]
\centering
\caption{\textbf{Ablation Study of the Proposed \acs{projection} on the nuScenes Dataset~\cite{nuscenes}.} Comparison of \ac{projection} and single uncertainty baseline. We adopt  ATE [m], $t_{\text{err}}$ [\%], and $r_{\text{err}}$ [$^\circ$/100m] as evaluation metrics.}
\label{tab:uncertainty_ablation_nusc}
\resizebox{1\linewidth}{!}{
\begin{tabular}{l | c | c |c c c | c c c}
\toprule

 & Training & Inference & \multicolumn{3}{c|}{Validation set} & \multicolumn{3}{c}{Test set} \\
\multirow{-2}{*}{\begin{minipage}[c]{1.5cm}\raggedright Methods\end{minipage}} & interval & interval & ATE & RPE$_{trans}$  & RPE$_{rot}$  & ATE & RPE$_{trans}$  & RPE$_{rot}$  \\
\midrule

Single uncertainty + DepthAnythingV2  & 2 & 1 & 0.808 & 0.032 & 0.052 & 0.893 & 0.039 & 0.055 \\
\acs{projection} + DepthAnythingV2 & 2 & 1 & \textbf{0.771} & \textbf{0.029} & \textbf{0.047} & \textbf{0.814} & \textbf{0.033} & \textbf{0.050} \\

\midrule
Single uncertainty + DepthAnythingV2  & 2 & 2 & 0.674 & 0.060 & 0.077 & 0.889 & 0.078 & 0.084 \\
\acs{projection} + DepthAnythingV2 & 2 & 2 & \textbf{0.629} & \textbf{0.054} & \textbf{0.068} & \textbf{0.744} & \textbf{0.065} & \textbf{0.077} \\

\bottomrule
\end{tabular}
}
\end{table}

\subsubsection{Masking comparison}
\Cref{fig:uncertainty_kitti_comparation} demonstrates that our \ac{projection} method is more robust than the self-discovered mask from~\cite{bian2021ijcv} and can detect complete areas where the static scene assumption is violated. Compared to single uncertainty predictions, which produce blurry masks around object contours and incorrectly assign high uncertainties to static regions, our method generates sharp boundaries that precisely identify assumption violations. Our approach learns low uncertainties for high-information regions such as road markings, while homogeneous areas remain less certain. This behavior provides meaningful supervision signals for PoseNet, focusing on distinctive features that do not violate the static scene assumption.

\begin{figure}[p]
\centering
\includegraphics[width=\textwidth]{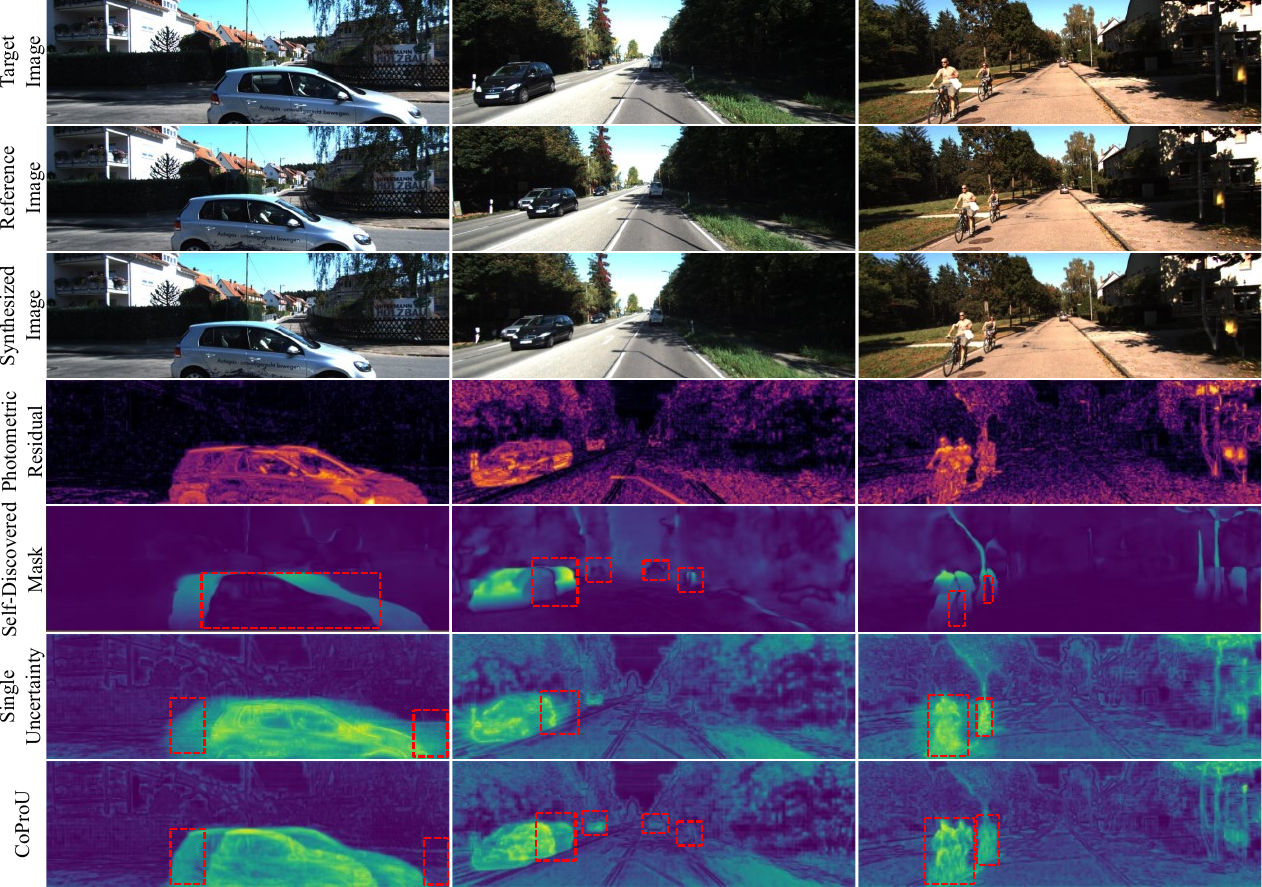}
\caption{\textbf{Comparison of Masking Mechanisms on KITTI~\cite{geiger2013vision}}. We compare the self-discovered dynamic mask from SC-Depth~\cite{bian2021ijcv}, the single uncertainty approach used in~\cite{yang2020d3vo,wimbauer2025anycam,wang2024self}, and our proposed \ac{projection}. The masks are predicted from both target and reference images. Our \ac{projection} accurately identifies and masks regions affected by dynamic objects with sharp boundaries. In contrast, the single uncertainty and self-discovered masks fail to fully cover dynamic regions and incorrectly mask static areas, producing blurry boundaries around object contours.}

\label{fig:uncertainty_kitti_comparation}

\end{figure}

\subsubsection{Visual analysis of uncertainty estimation} \Cref{fig:further_uncertainty} shows visualizations of our \ac{projection}. Our method successfully detects dynamic objects (e.g., moving cars) and distinguishes them from static objects (e.g., parked cars). Additionally, it assigns low uncertainties to areas with meaningful information such as road markings, manhole covers, and building edges, while being uncertain in homogeneous regions where even humans would struggle to use them as references for localization and pose estimation.

\begin{figure}[!ht]
\centering
    \centering
    \includegraphics[width=\textwidth]{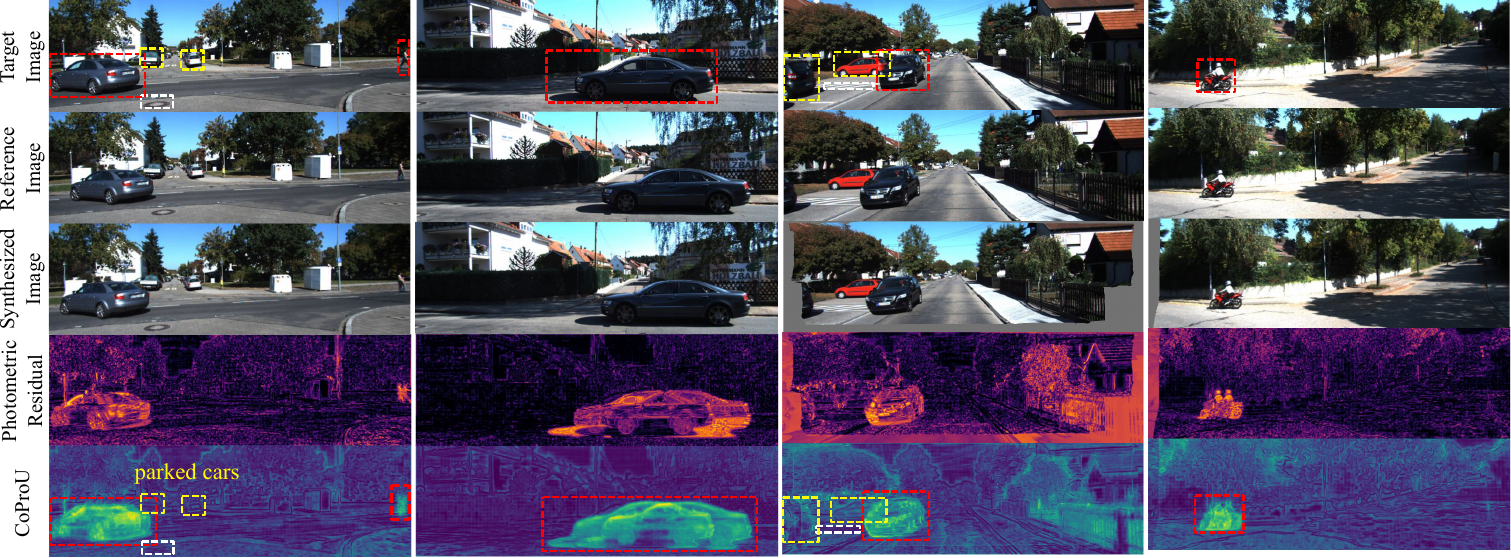}
     \text{(a) KITTI dataset~\cite{geiger2013vision}}
     
    \vspace{1\baselineskip}
    \includegraphics[width=\textwidth]{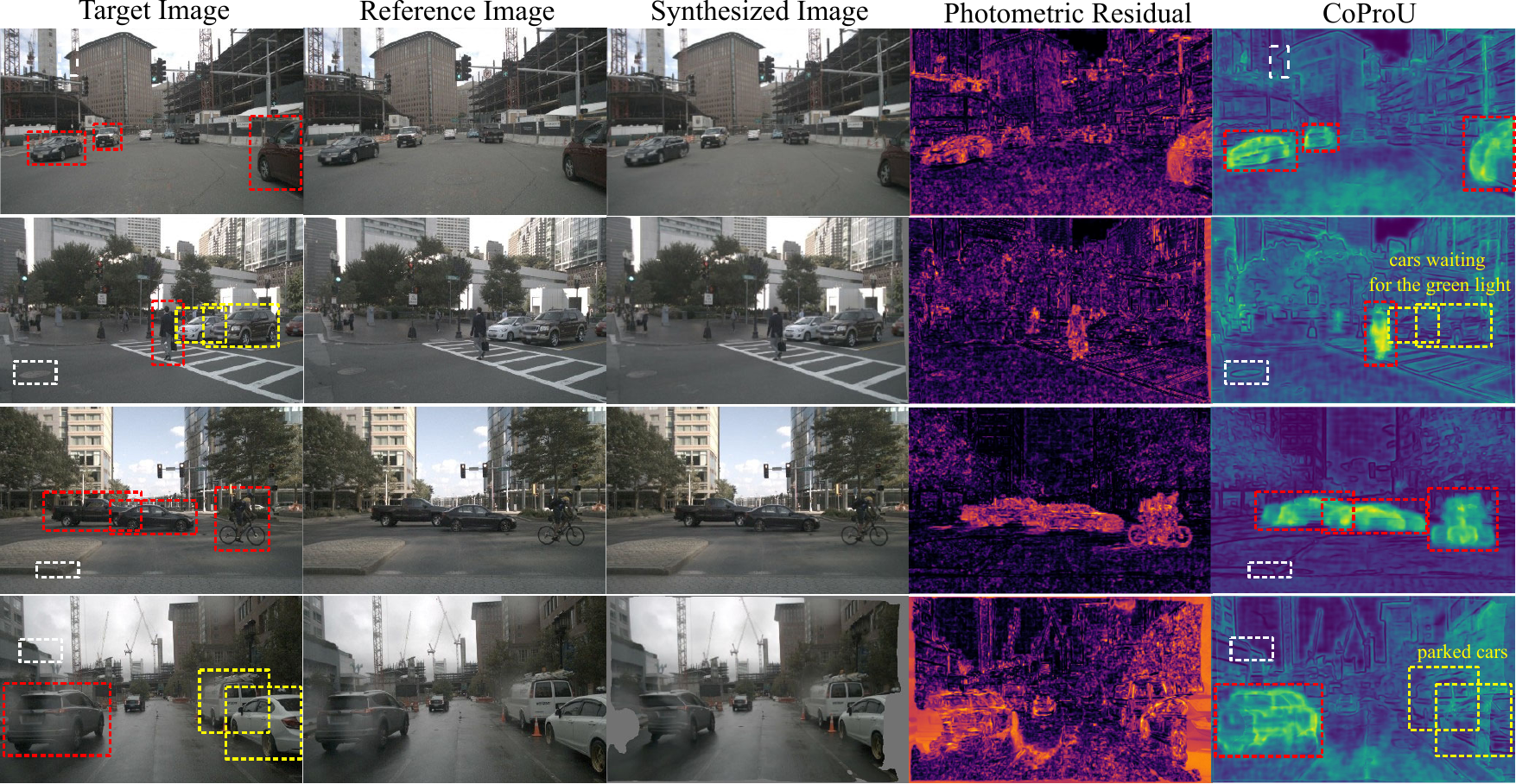}
    \text{(b) nuScenes dataset~\cite{nuscenes}}

\caption{\textbf{Further Uncertainty Visualization}. Our \ac{projection} effectively detects dynamic objects (red boxes) by assigning high uncertainties, while static objects (yellow boxes) are correctly identified as certain features. Informative features such as road markings and structural edges (white boxes) are assigned low uncertainties, demonstrating the method's ability to distinguish between reliable and unreliable regions for pose estimation.}
\label{fig:further_uncertainty}
\end{figure}

\subsubsection{Occlusion analysis}
\Cref{fig:uncertainty_occlusion} demonstrates our the capability of the proposed \ac{method} to assign high uncertainties to occluded regions where the static scene assumption is violated.

\begin{figure}[!ht]
\centering
    \centering
    \includegraphics[width=\textwidth]{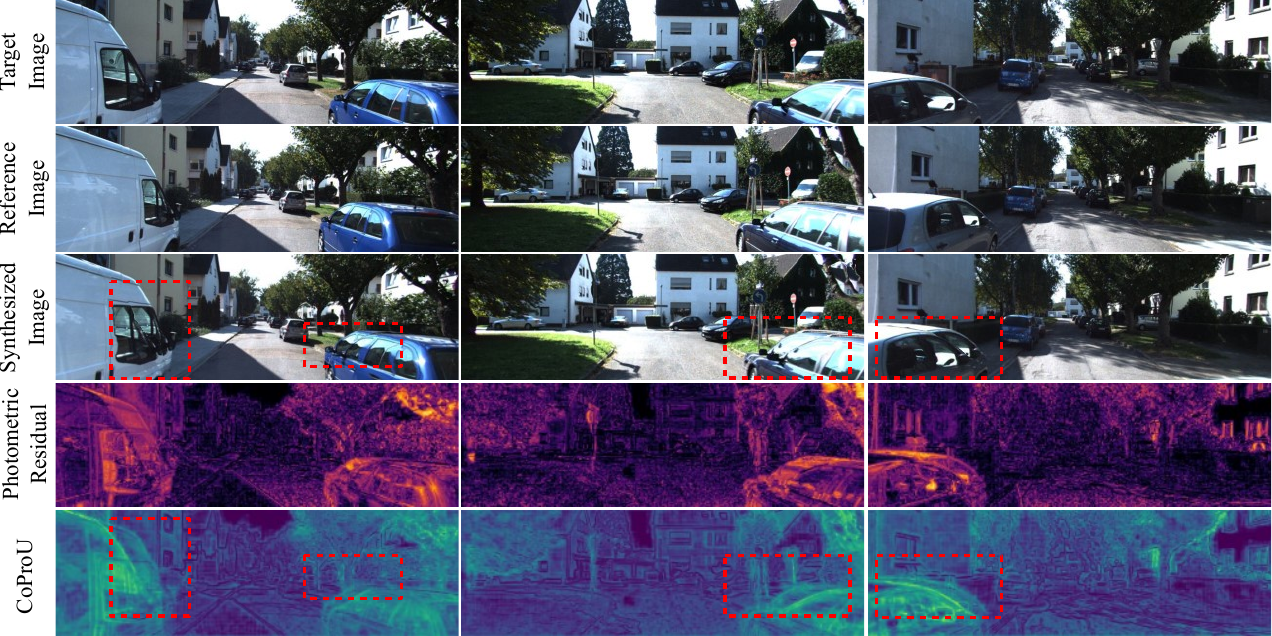}
    \text{(a) KITTI dataset~\cite{geiger2013vision}}
    
    \vspace{1\baselineskip}
    \includegraphics[width=\textwidth]{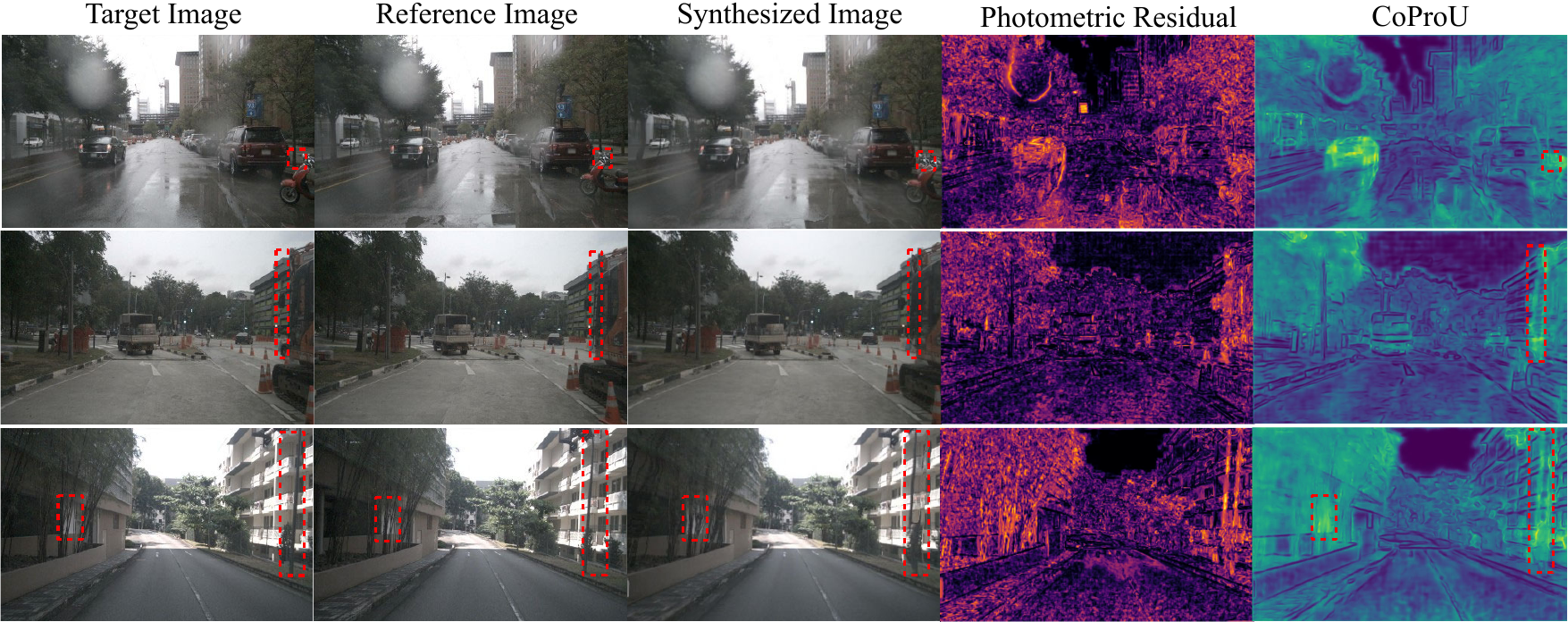}
    \text{(b) nuScenes dataset~\cite{nuscenes}}
  
\caption{\textbf{Uncertainty Visualization in Occlusion Scenarios}. Our \ac{projection} handles occlusion scenarios across both KITTI~\cite{geiger2013vision} and nuScenes~\cite{nuscenes} datasets, demonstrating robust uncertainty estimation when objects are partially occluded or when visibility is limited.}
\label{fig:uncertainty_occlusion}
\end{figure}

}{}

%
%
%
%

\newpage
\bibliographystyle{splncs04}
\bibliography{027-main.bib}

\end{document}